\documentclass{report}
\usepackage{setspace}

\usepackage{amssymb,graphicx,color}
\usepackage{amsfonts}
\usepackage{latexsym}
\usepackage{a4wide}
\usepackage{amsmath}
\usepackage{commath}
\graphicspath{ {./images/} }
\usepackage{placeins}

\title{{\vspace{-14em} \includegraphics[scale=0.4]{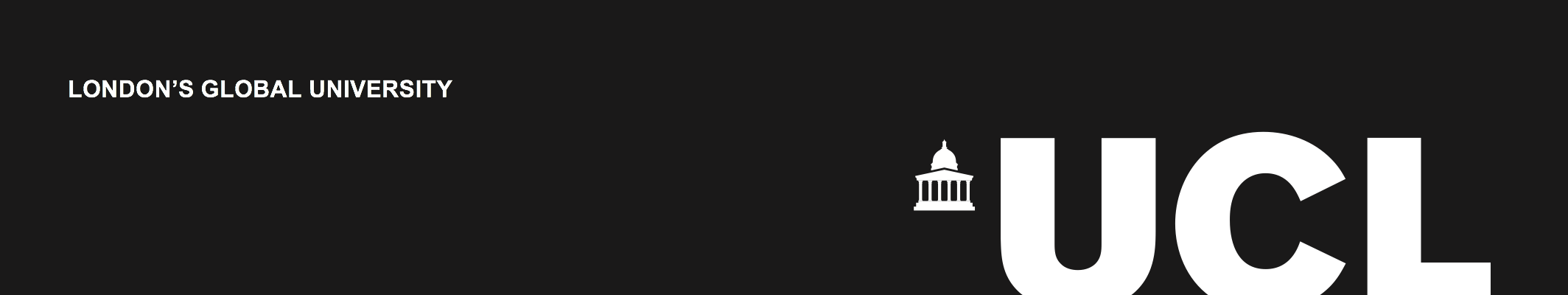}}\\
{{\Huge Recognition and Synthesis of Object Transport Motion}}\\
}
\date{Submission date: 29 April 2019}
\author{Connor Daly\thanks{
{\bf Disclaimer:}
This report is submitted as part requirement for the Computer Science MEng at UCL. It is
substantially the result of my own work except where explicitly indicated in the text.
 The report may be freely copied and distributed provided the source is explicitly acknowledged}
\\ \\
MEng Computer Science\\ \\
Yuzuko Nakamura and Tobias Ritschel}

\begin{document}
 
\onehalfspacing
\maketitle
\begin{abstract}
Deep learning typically requires vast numbers of training examples in order to be used successfully. Conversely, motion capture data is often expensive to generate, requiring specialist equipment, along with actors to generate the prescribed motions, meaning that motion capture datasets tend to be relatively small. Motion capture data does however provide a rich source of information that is becoming increasingly useful in a wide variety of applications, from gesture recognition in human-robot interaction, to data driven animation. 

This project illustrates how deep convolutional networks can be used, alongside specialized data augmentation techniques, on a small motion capture dataset to learn detailed information from sequences of a specific type of motion (object transport). The project shows how these same augmentation techniques can be scaled up for use in the more complex task of motion synthesis. 

By exploring recent developments in the concept of Generative Adversarial Models (GANs), specifically the Wasserstein GAN, this project outlines a model that is able to successfully generate lifelike object transportation motions, with the generated samples displaying varying styles and transport strategies.
\end{abstract}

\renewcommand{\abstractname}{Acknowledgements}
\begin{abstract}
I would like to express my sincere gratitude to my supervisors, Yuzuko Nakamura and Tobias Ritschel for giving me the opportunity to work on this interesting project. Yuzi's encouragement and advice was always helpful, while Tobias's technical expertise was instrumental in keeping the project on track.

\end{abstract}
\tableofcontents
\setcounter{page}{1}

\chapter{Introduction}

A significant proportion of how we communicate with others is reliant on the ability to successfully recognize gestures and motions. Indeed, how we move is an incredibly rich source of information. It has been shown, for example, that a person’s gait is unique enough to allow for individual recognition systems to be built, reliant solely on the the silhouette of the person walking \cite{han}. It is unsurprising therefore that given the richness of motion data, and its importance in human interaction, there remains a high interest in both the analysis and synthesis of complex human movement. 

A shift in the paradigm of how robots are used has led to a spike in the research of automated gesture recognition. Instead of being shuttered away in factories with the `lights out', robots are increasingly used in a more collaborative setting, working alongside humans. The ability to recognize human activities and gestures is vital for intuitive human-robot interaction. In addition to this, activity recognition systems have potential for use in a wide variety of applications today, such as security, health-care and even energy conservation in buildings \cite{ngu}. Detecting if an elderly patient has fallen may be critical in getting them the treatment they need, while the ability to continuously recognize the activities and intents of individuals at secure locations such as airports is of particular interest to law enforcement. 

Although for many of these purposes, the focus is on recognizing what people are doing (activity recognition) rather than recognizing how they are performing the particular activity, within this report we focus on learning nuanced information from one particular set of motion, namely object transportation. Classifying how a person transports an object may be particularly useful, as for example, detecting if an individual is struggling with a heavy object is applicable to numerous applications. To this end, the report describes how deep learning can be used to identify characteristics of the object an individual is transporting.

With regard to motion generation, synthesizing realistic lifelike human movement is an often complex and difficult process. Although recent years have seen huge leaps in graphics and rendering technologies allowing for the creation of photo-realistic character models,
using current industry standard techniques, generating realistic human motion remains a laborious task, with animators taking weeks to produce minutes worth of footage using key-framing techniques \cite{las}. Motion capture techniques can greatly facilitate the process, essentially allowing animators to `trace' the movements of an actor and fill in character details within the outline. However, direct application of motion capture is often limiting in the regard that the character must follow exactly the motion path performed by the actor. For applications such as video games or virtual environments, the lack of richness in terms of motions available to a character may lead to a decrease in the immersion of the experience. Thus, a means of quickly generating creative non repetitive motions is needed. 

To overcome this limited motion availability, by interpolating over a base set of motions, it is possible to generate a continuous space of motion \cite{kov}. Within this report we outline techniques that lead to the generation of the continuous space of one particular motion, object transport. This space defines motion sequences varying in style and transport strategy. 

As outlined later in the report, approaches that make use of machine learning techniques show strong promise in tackling both the task of generating diverse realistic motions and the task of motion recognition. Specifically, we are interested in highlighting the suitability of deep convolutional models for both motion sequence analysis and synthesis.

\subsection{Report Outline} 

Within this report, for both motion recognition and motion synthesis we make use of a dataset comprised of 13 individuals moving a bowl from one side of a room the other. 

With regard to motion recognition, this report outlines how deep learning techniques can be used to learn nuanced information from an object transportation motion sequence. Each sample within the dataset varies in terms of motion style and transport strategy. This variation is due, in large part, to the characteristics of the object the individual is transporting. From a given motion sequence, the proposed neural architecture is able to classify if the individual is trying to keep the object balanced, if the object is heavy or light, and furthermore, classify the given motion into a particular subclass of motion, or strategy. 

The latter half of the report diverges from motion recognition, and focuses on the generation of motion sequences using Generative Adversarial Networks (GANs), wherein two networks are trained against each other, with one network attempting to produce motion sequence paths convincing enough to fool the other network. The report explores recent developments in the GAN concept, and shows how such networks can be used to synthesize lifelike object transport motion sequences. The variation within the used dataset allows the model to generate a continuous motion space from a base set of motion strategies and styles. Beyond this, the report also shows how different styles of motion can be generated when the network is conditioned by factors such as the heaviness of the bowl, thereby refining the types of sequence generated by the model. 

Finally, the report shows how data augmentation methods can be applied within the context of motion sequences to allow for the use of deep learning techniques, which typically require a vast training dataset, on a relatively small set of examples with a large feature-space. 

\begin{figure}[h]
	\centering
	
		\makebox[\textwidth][c]{\includegraphics[width=1.35\textwidth]{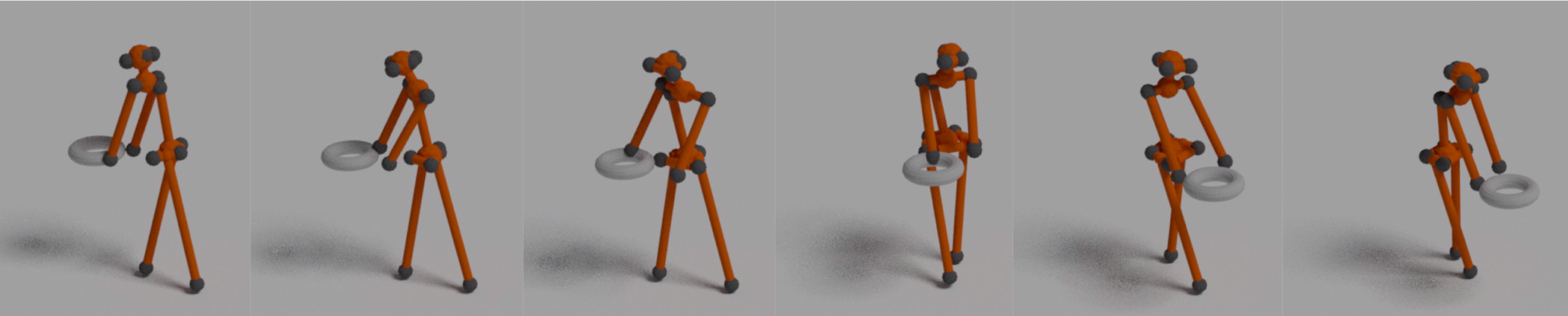}}
		
		\caption{Motion sequence generated by WGAN-GP model outlined in the report}

\end{figure}
\FloatBarrier

\chapter{Related Work:}

\section{Motion Recognition}

Human activity recognition has applications across many diverse domains, such as sports video analysis, surveillance and robotics. 
Humans, as expected, are incredibly adept at recognizing what their companions are up to. Observations from a study in 1973 \cite{json}, suggest that humans are able to recognize actions from the motion of 8 light displays attached to the human body. 

Motion capture systems can provide a similar 'skeleton' representation of the body, giving the three-dimensional coordinate locations of markers fixed to the body. Similar joint representations can be extracted fairly precisely via depth sensors such as the Microsoft Kinect. However, learning to recognize human activity from this representation remains an open problem within the field of computer vision.

An early approach to this problem involves learning 'templates' that represent a class of motion,with  each template being a matrix with rows corresponding with joint positions, and columns corresponding with time \cite{temp}. When classifying new motions, a K nearest neighbour classifier is used, matching joint positions to its nearest template. This method is susceptible to producing large temporal misalignment however, particularly so for periodic actions such as hand waving. 

A different approach used by Xia et al. \cite{xia}  involves dividing the 3D space where the action takes place into a number of bins using a spherical coordinate system.  For each frame of motion, joints are cast into certain bins with a Gaussian weight function, forming a histogram. Following this, these histogram sequences are then clustered into k cluster words. Thus, each frame of motion will reside within a particular cluster word, and each action corresponds with a sequence of such words.  Following this, a Hidden Markov Model (HMM) is used to model the temporal evolution of the postures. 

 Wang et al. \cite{wang} make use of a Fourier Temporal Pyramid (FTP) to model the evolution of pose over time. For a complex articulated structure, the FTP models how the motion of individual parts are correlated. This can be seen to be a far more natural way of describing an action than simply focusing on the change in trajectory for all joints. For example, when walking, as the left arm swings forward, the right swings back. Further to this, the authors of this paper describe 'actionlets', that are defined by a particular cluster of features given by a subset of joints.  One human action is therefore an actionlet ensemble, or a linear combination of actionlets. For a particular action, the discriminative weights of its component actionlets are learned through kernel methods. The results of these are then ensembled to give the predicted action. 

Recently, deep learning has come to dominate the approaches to tackling the challenge. Cho and Chen \cite{cho} use a Multi Layer Perceptron (MLP) to classify each frame of a motion sequence, however this network structure is unable to explore temporal dependencies and look at how poses evolve over time within certain actions. In contrast to MLPs, Recurrent Neural Networks (RNN) with LSTM (Long Short Term Memory) were designed with temporal data in mind  \cite{gers}. LSTMs recognize patterns across time using historical context. For example, with respect to machine translation, given a sentence, how an LSTM will translate a particular term is dependent on all the words that precede that term in the sentence. 
Consequently many of the best preforming machine translation models making use of such networks \cite{bert}.

Given this suitability for temporal learning, it is little wonder that approaches  that make use of LSTM frameworks show promise with regard to the analysis of motion sequence data. In a similar fashion to the actionlets defined over a subsection of joints, Du et al. \cite{Du} take a novel approach wherein they divide the joints of the skeleton representation into 5 clusters of joints. For a particular motion sequence, each group is fed into a separate bidirectional RNN with LSTM. The results from this are then concatenated to form 2 clusters representing the upper and lower body, which are then run through another 2 separate RNNs, before the concatenated results are then run through the final RNN layer.  On Berkeley's MHAD dataset, consisting of 11 motion classes, the performance of this model beats all other approaches, including those using  HMMs. 

Unlike LSTMs which are often used in the context of times series data, Convolutional Neural Networks (CNNs) are more often applied to the field of image processing and recognition. With relation to images, CNNs slide filters across the pixel input of the image, thereby learning the spatial relationship between local image features. However recently, there has been an upsurge in research showing how the same principles can be applied to learning the relationship between temporal features, highlighted, for example, by Google's  `Wavenet' paper \cite{wav} showing how a CNN can be used to learn from streams of audio data. In the context of temporal data, whereas LSTMs rely on all preceding time-steps when learning from a sequence, CNNs make use of localized information. Given this localized context space training a CNN tends to be less computationally expensive than an LSTM used for the same task.

The potential for convolutional models to learn from motion sequence data is shown by Du et al. \cite{du2}. By using a similar hierarchical structure to their previous RNN-LSTM model, they are able to achieve the same accuracy on the MHAD dataset as they do with their previous model.

Taking motivation from these results, we aim to develop a CNN that makes use of similar hierarchical joint clustering and apply it to a new dataset. We aim to test the efficacy of convolutional models in discriminating nuanced information from object transport motion sequences, such as the weight of the object being carried and if the individual is balancing the object, on a small dataset.

\section{Motion Synthesis}
With motion capture, aside from the growing interest in analysis and classification of actions, there has also been an upsurge of interest in synthesis of motion. There are currently two main approaches to the task of motion synthesis: physics based, and data driven.  
 
 \subsection{ Physics Based Approaches:}

 The physics based approach often involves modelling the task at hand as a torque minimization problem. The virtual actor is first parameterized by biophysical constraints. The synthesis model is then optimized over these constraints, such that the virtual actor completes the prescribed task with minimal torque exertion. This has seen some notable success. A notable proponent of this technique, C. Liu, has used this approach to generate realistic movement in not just humanoid actors, but also in swimming creatures \cite{lu1}. Of particular interest is her application of the technique to generating realistic hand manipulation animations across a wide variety of tasks \cite{lu2}, given the object to be manipulated, an initial pose and the intended end result. 
 
 The rise of reinforcement learning has also led to new approaches in physics based methods. Again the virtual actor's bio-physical constraints must be preset, however unlike the previous method described, reinforcement learning techniques are optimized over some higher level, task specific metric, for example with regard to locomotion, this could be how far the virtual actor is able to move prior to falling over. Google's Deepmind has recently shown that policy optimization techniques can be used to train virtual actors to be able to successfully navigate simulated `parkour' environments \cite{Silver}. Similar policy optimization techniques were successfully employed by Clegg et al for the synthesis of human dressing motion \cite{Clegg}.
 
 For all physics based approaches however, robustly parameterizing the motion task at hand, and accurately representing the bio-physical constraints of the virtual actor, and the physics of the world around them, in order to drive these models remains a complex and difficult task. Furthermore, reinforcement learning requires not only huge amounts of precomputation, which exponentially increases with the number of actions available to the virtual actor, whilst the results are also heavily dependent on how optimization problem is framed and may lead to motion that is energy efficient but not necessarily lifelike. 
 
 \subsection{Data driven approaches}

\textbf{Linear interpolation and kernel based methods:}

In contrast to physics based techniques, data driven motion synthesis techniques draw on motion capture samples taken from a preexisting motion database. The model then interpolates on these base samples and generates a continuous space of motion.

The simplest form of this approach was first proposed by Hahn and Wiley \cite{wiles} which involves simply taking a linear interpolation between multiple motions in the database. This has been shown to work well for a small set of base motions, however, the computational cost increases exponentially as the number of motion samples increases. 

Accordingly, kernel based methods go some way in reducing the computational cost. Radial-basis functions (RBFs) have been shown to be effective in interpolating multiple motions of the same type \cite{Rose}. However, these methods lack the means to handle noise and variance, and thus have a tendency to over-fit data. 

In contrast, Mukai and Kuriyama \cite{Mukai} have shown that Gaussian Processes do not suffer from these issues. Wang et al. \cite{Wang} further build on this by applying a Gaussian Process Latent Variable (GPLV) model across time-series motion data to learn the posture in the next frame given the previous frames.
Unfortunately, with these methods, the number of motions which can be interpolated is also limited, this time by the large memory cost.

\textbf{Deep Learning Approaches:}

Neural Networks can be thought of as powerful function approximators. Within the context of motion synthesis, the weights of a trained network are able to represent the components of an individual motion. When an input is passed into the network, it is transformed by these weights, essentially resulting in a weighted sum of each of these components. The ability of a neural network to represent the motion space within its weights, rather than a stored motion database, leads to low memory costs. Furthermore, besides the initial cost of training the network, generating motion from the trained network is computationally inexpensive, allowing for the near instantaneous creation of varied and interesting movement.

Taylor et al. \cite{Taylor} make use of conditional Restricted Boltzmann Machines for synthesizing gait animations. Fragkiadaki et al. \cite{Frag} apply an Encoder-Recurrent-Decoder network, a form of RNN, to first learn the features of the motion manifold along with temporal dynamics,  in order to produce smooth motion interpolations. This method is shown to be much more scalable and significantly more run-time efficient than its linear or kernel-based counterparts. 

Holden et al. show how a convolutional auto-encoder can be used to reduce the dimensionality of the motion manifold. By combining a Multi-layer Perceptron (MLP) with the trained auto-encoder, this can then be used to generate motions belonging to a particular domain, for example locomotion, or punching or kicking \cite{hol}.

Away from motion synthesis, one of the most recent innovations in the area of generative models is the Generative Adversarial Network (GAN), pioneered by Goodfellow et al. \cite{good} Using this model, they are able to generate incredible `fake' face images using the Toronto Face Database. Since the release of the paper in 2014, GANs have been used to achieve incredible results with regard to image generation, however the use of GANs outside image generation is somewhat limited. Vondrick et al. have shown some promising results for the use of GANs in video frame prediction \cite{vids}. In this project, we hope to leverage the strength of GANs to uncover the underlying data distribution of motion transport, and from this generate new motions.

\subsection{Summary}

This chapter first introduces the idea that human motion can be meaningfully represented by a `skeleton' of points in space. The chapter then gives an overview of the traditional approaches to tackling motion recognition, such as through the use of Hidden Markov Models, before mentioning the rise of deep learning within this problem space. 
Following this section, the chapter also introduces the fact that there are two main categories of motion synthesis model: physics based, and data driven. For each category, the chapter also details specific approaches and highlights their respective advantages and disadvantages.

\chapter{Background Information}

\section{Convolutional Neural Network}

The efficacy of Convolutional Neural Networks( CNNs) in the field of image classification is one of the key driving forces behind the recent resurgence in deep learning. However besides image processing, CNNs have also been adapted to process many other kinds of data, recently seeing success in the field of Natural Language Processing and even music, as stated in chapter 2.

CNNs take inspiration from the way we process information with the visual cortex, the region of neurons in the brain sensitive to particular areas of the visual field. It has been shown that particular neurons in this region only respond when the eyes are shown images of edges with a particular orientation such as vertical or diagonal \cite{hub}. These neurons are arranged in columns. At a high level, it can be said that CNNs work in a similar way, with particular parts of the architecture being responsible for recognizing certain features of the input data, with early layers extracting low level representations such as edge orientations, and later layers building on these to extract more complex representations.

With the context of this report in mind, we shall use the example of a CNN processing some time sequential data, such as a motion sequence. Using this example we have 2 dimensions as input: the first being time; and the other being the feature recorded at the time-step. For this particular example, we will say there are 32 time steps and 3 features, therefore a 32$\times$3 input.  

Within a CNN, there are several core operations: convolutions and pooling. 

A convolutional layer slides, or convolves, filters along the input data. In the context of image processing convolve across 2 dimensions, across and down pixels of a particular image, learning the spatial relationships between pixels. With time data however, filters convolve only over the time axis, performing a 1D convolution thereby learning the relationship between time-steps in the sequence. 

As the filter convolves over the sequence, the weights within the filter are multiplied point-wise over the features they are covering. Within the example, if the filter size is of size 3 time-steps, the filter is a matrix of 3$\times$3 weights. We then take the Hadamard product of the filter weights with the time-steps that the filter is covering. The values of the resulting matrix are then summed to give a scalar. If the stride of the convolution is 1, the filter performs this operation over every set of 3 sequential time steps, otherwise the filter jumps k time-steps, where k is the stride factor. 

Prior to passing the filter over the input data, we append 2 padding time-steps of zeros before the first time-step, and 2 after the last. Thus, if k is 1, convolving 1 filter over the input results in a 1D array of size 32. If k is 2, the array will be of size 16. This array is then passed through a non-linear function such as the sigmoid function, or tanh. 

\begin{figure}[htbp!]
	\centering

		\includegraphics[scale=0.8]{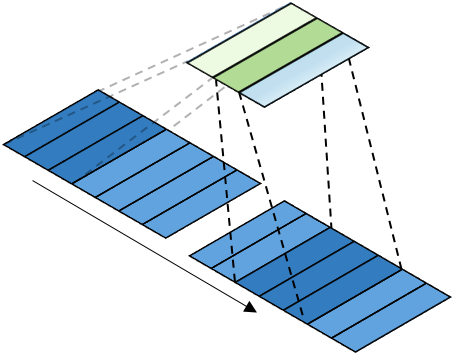}
		\caption{1D convolution with stride length 2}

\end{figure}
\FloatBarrier

The area of the datapoint on which the filter is positioned is the receptive field. A large filter size gives a large receptive field, allowing the filter to extract a feature that is spread across the datapoint, or spread across a long time-span in our case. However a large filter has increased computational complexity. To increase the span of the receptive field without this overhead we can use a dilated convolution, wherein we pad d rows of zeros around every row of the filter, where d is the dilation factor. Thus instead of learning relationships between features in 3 successive time-steps, the filter now learns the relationship between features at 3 points in time within the sequence, each separated by d time-steps. 

\begin{figure}[htbp!]
	\centering

		\includegraphics[scale=0.8]{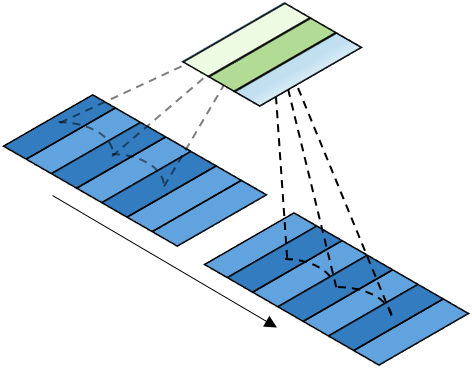}
		\caption{dilated convolution of spacing 1 along time axis}

\end{figure}
\FloatBarrier

A convolutional layer may contain several filters. If there are 3 filters in the first convolution, and k is 1, the output of this layer will have a size $32\times3$. As stated, deeper layers within the CNN extract high level representations from low level representations extracted by earlier layers. Adding more filters to a layer can give a richer representation of the input data to the network, however the increased number of filters come with added computation complexity. In order to decrease this complexity, CNNs include a downsampling layer, such as a max pooling layer. 

This layer resembles the convolutional layer somewhat in that a pooling 'window' is slid over the layers input. Within our example, using a pooling window of size 3, we take the maximum value from the 3 time-steps the window covers. Besides reduced complexity, briefly going back to edges and curves within image classification, the intuitive idea behind pooling is that the exact location of the edge when found is not as important as its rough position with respect to other regions in the image. This loss of granularity allows the network to better generalize and reduces the chance of overfitting. With respect to time sequences, this can help the network effectively learn from data that is not perfectly aligned. Pooling windows of size 2, with a stride of 2 are often used, reducing the input tensor size by half.

The training process of a CNN is similar to that of other neural networks. When used for classification, the final dense layer of the network is typically a softmax function. This forces the output of the network to represent a distribution.  Each node in the output layer represents a certain class, thus the sum of the values output by the last layer is 1. With these probabilities and the ground truth, an error is computed, which is then back-propagated through each layer in the network. Stochastic gradient descent (or another optimizer function) is used to adjust the weights such that this error is minimized as training goes on. 

Typically the input data to CNNs is normalized in a way that ensures the network input follows a normal distribution, with a mean of 0 and unitary variance. This is to prevent early saturation of non-linear functions used after each convolutional layer. Although this is sufficient to stave off gradient saturation in early layers, issues can arise deeper within the network as the distribution of the activation outputs constantly shifts throughout training. This can slow the training process as layers adapt to a new distribution during each training step, or even a complete stop on the learning process as the gradient vanishes.  

Batch normalization forces input to the subsequent layer to have approximately the same distribution as the previous training step, by using the mean and variance values computed from the previous batch.

\subsection{Generative Adversarial Networks}

In recent years, the power of Generative Neural Networks (GANs) has captured headlines. From  `AI artists' \cite{chris} generating beautiful works of art, to giving even the most awkward among us the `ability' to dance \cite{chan}, GANs are capable of modelling incredibly complex data distributions. 

Our focus for the next stage of the project is on the use of GANs to be able to approximate the underlying distribution behind object transportation. 

Generative Adversarial Networks (GANs) are in fact made up of two distinct networks, a generator network, and a discriminator network. The generator network maps a noise vector to the input space, which in our case is the 32x48 axis coordinates from each marker at each frame within a sample. The discriminator network will receive either a true data sample, or one generated by the generator network, and must distinguish between each of them. 

Many types of generative models, such as Variational Auto-Encoders rely on minimizing the Kullback-Leibler (KL) divergence, a metric of how close two distributions, for example p and q, are from each other.

$$D_{KL}(p||q) = \int_{x} p(x)\log\frac{p(x)}{q(x)} dx$$

$D_{KL}$ is 0 when p(x) is always equal to q(x). It is important to note however that KL divergence is asymmetric. Looking at the formula, when p(x) is near 0, but q(x) is significantly higher, this disparity is ignored according to the KL metric. This is problematic when attempting to measure the distance between two equally important distributions, such as that of the generator and that of the true data distribution. 

Accordingly, to combat this issue, GAN training is equivalent to the minimization of Jensen-Shannon divergence, a symmetric measure of distribution closeness.

$$D_{JS}(p||q) = \frac{1}{2}D_{KL}(p||\frac{p+q}{2}) +  \frac{1}{2}D_{KL}(q||\frac{p+q}{2})$$
\newline

The generator is trained so that it is able to fool the discriminator. The two networks are trained alternately, resulting in a min max, or 'zero-sum' game across the following objective:

$$\min_{G} \max_{D} \mathbf{E}_{x\sim \mathbf{P}_{r}}[\log(D(x))] + \mathbf{E}_{\bar{x}\sim\mathbf{P}_{g} }[\log(1-D(\bar{x}))] $$
\newline
Where $\mathbf{P}_{r}$ is the distribution of the observed data, and $\mathbf{P}_{g}$ is the generator distribution, defined by $\bar{\mathbf{x}} = G(z)$, with z being the noise input, sampled from a simple distribution, such as uniform or Gaussian.

The generator and discriminator should be trained until they reach a Nash equilibrium, the point at which, each network cannot decrease its own cost without modifying the other's parameters. During training we must take care to balance the relative strength of both discriminator and generator network. 

GANs are highly sensitive to hyper-parameter choice. Furthermore, the generator network is susceptible to to diminished gradients. Early in training, when G is weak, or when D grows too strong, D rejects samples produced with high confidence, due to the clear difference from training data. This leads to the saturation of $\log(1-D(\bar{x}))$. The original GAN paper suggests an alternative loss function to counter this issue, however Arjovsky shows this leads to highly unstable updates \cite{arj}. 

\subsection{DCGAN}
In the original GAN paper, the discriminator and generator networks are Multi-Layer-Perceptrons (MLPs), made up purely of fully connected dense layers. Although this structure performs well on the MNIST dataset, as well as on the Toronto Face Database, it has proven difficult to train across other datasets, with results often appearing noisy and incomprehensible \cite{alec}. The Deep Convolutional GANs (DCGANs), makes use of the strengths of CNNs in learning the spatial relationships between different features of the input, by replacing the dense layers in the original model's generator and discriminator network with convolutional layers. The DCGAN structure has been shown to be relatively stable to train on image data, whilst capable of producing state-of-the-art results \cite{bang}. The discriminator network, in contrast with our network developed for the preliminary experiments, all maxpooling operations are replaced with strided convolutions, thereby allowing the model to learn its own downsampling. In the generator network, transforming input from the dimensions of the noise space, to the dimensions of the data is done through up-sampling layers.

\subsection{WGAN}

The Wasserstein GAN (WGAN) proposed in 2017 by Arjovsky et al. \cite{wgan} replaces the loss function of the original GAN paper, with Wasserstein distance, or 'earth-movers distance' (EM). The EM is a measure of the effort required, if we imagine each distribution as a particular pile of earth, to transform one probability distribution pile to match the other.

\begin{figure}[htbp!]
\centering
\includegraphics[scale=0.5]{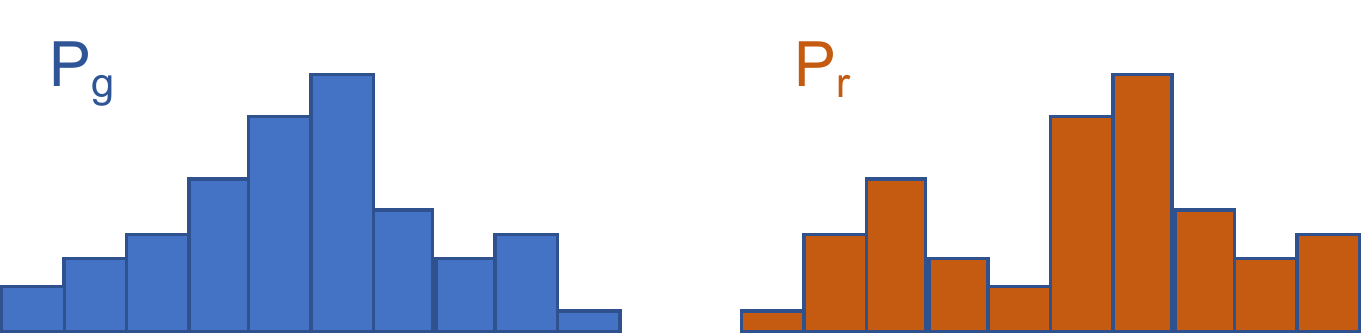}
\caption{2 distributions $P_{g}$ and $P_{r}$}
\end{figure}
\FloatBarrier
The formula calculating this distance between two distributions $P_{r}$ and $P_{g}$ with possible states x and y respectively is described below:

$$W(P_{r},P_{g}) = inf_{\gamma\sim\prod(P_{r},P_{g})} \mathbf{E}_{x,y\sim\gamma}\norm{x-y}$$

Wherein $\prod(P_{r},P_{g})$ is the set of all joint probability distributions of $P_{r}$ and $P_{g}$.

A particular $\gamma$ describes 1 possible transport plan for moving a percentage of earth from a point at x to a point at y such that the distribution over x follows that of y.  Thus $\sum_{x}\gamma(x,y) = p_{g}(y)$, or after moving the planned amount of earth from each possible state x to its make its target y, we have exactly the distribution over y. 
Going from x to y, the distance travelled is $\norm{x-y}$, the amount of earth moved is $\gamma(x,y)$ therefore the cost of going from x to y is $\gamma(x,y)\cdot\norm{x-y}$, and therefore the expected cost for all pairs of x y is denoted as: $$\sum_{x,y}\gamma(x,y)\norm{x-y}=\mathbf{E}_{x,y\sim\gamma}\norm{x-y}$$

The Wasserstein distance is the transport plan with the lowest associated cost, denoted by $infimum$, or greatest lower bound.

To see how Wasserstein distance can provide a better loss metric within the application of GANs we look at 2 simple distributions, P and Q, defined by a parameter $\theta$ where within P, in all joint distributions over x and y, x=0 and $y\sim U(0,1)$, and within Q, in all joint distributions over x and y,  $0 \leq\theta \leq 1 $, x$ = \theta$, and $y\sim U(0,1)$.

\begin{figure}[htbp!]
\centering
\includegraphics[scale=0.6]{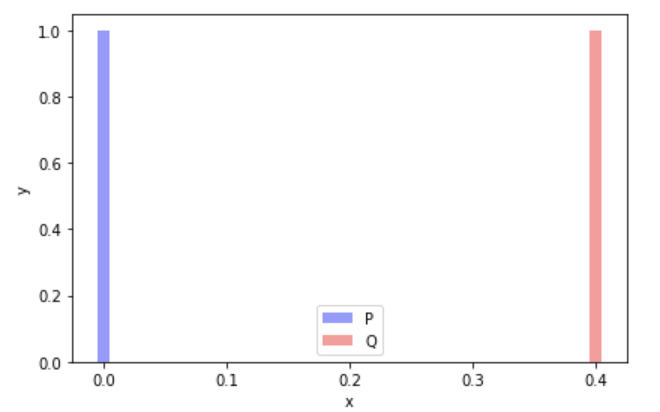}
\caption{When $\theta$ is not 0, there is no overlap in distributions P and Q}
\end{figure}
\FloatBarrier

The two distributions fully overlap when $\theta=0$. In this case the JS divergence, $D_{JS}(P||Q) = 0$. However when $\theta$ shifts from 0, the JS divergence jumps to $log2$
In contrast, for both cases of $\theta$, the Wasserstein distance is equal to $\norm{\theta}$. Showing Wasserstein distance and JS divergence as a function of theta, we can see that Wasserstein distance for these two distributions is continuous and provides a usable gradient everywhere, unlike JS divergence.

\begin{figure}[htbp!]
\centering
\includegraphics[width=1\textwidth]{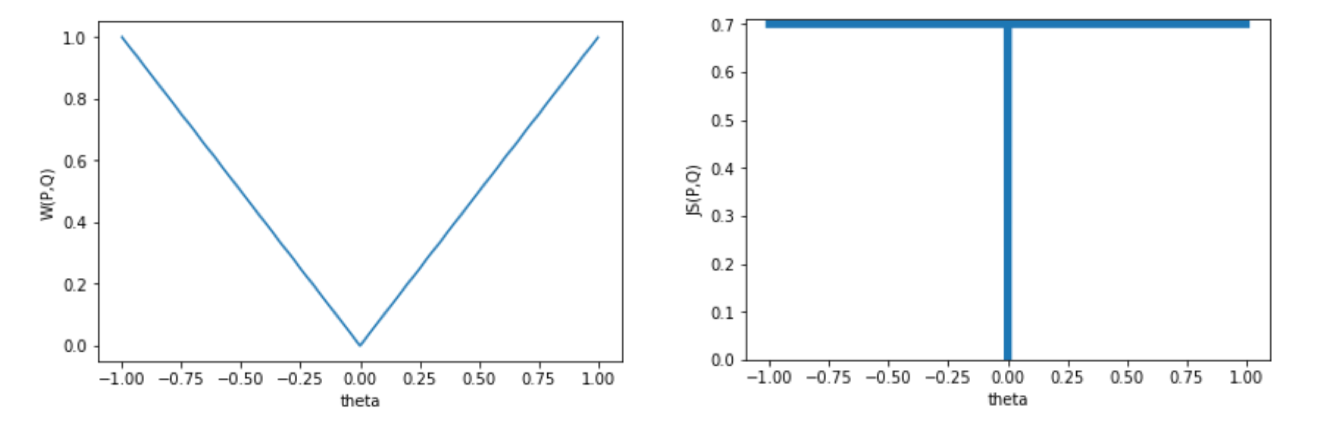}
\caption{Wasserstein distance between P and Q as a function of theta on the left, and JS divergence between P and Q as a function of theta on the right}
\label{fig:theta}
\end{figure}
\FloatBarrier
However, going back to our original distributions of $P_{r}$ and $P_{g}$ it is intractable to find all the possible joint distributions of $\prod (P_{r},P_{g})$ to compute $inf_{\gamma\sim\prod(P_{r}, P_{g})}$.

Consequently, Wasserstein distance is reformulated using the Kantorovich-Rubinstein duality to:

$$W(P_{r},P_{g})= sup_{\norm{f}_{L}\leq 1} \mathbf{E}_{x\sim P_{r}}[f(x)] - \mathbf{E}_{x\sim P_{g}}[f(x)]$$

where $sup$ denotes suprenum, or maximum value, and f is a 1-lipschitz function that follows the constraint:

$$|f(x_{1})-f(x_{2})|\leq|x_{1}-x_{2}|$$

Within WGAN, this function is learned by a neural network, which replaces the discriminator function in the original GAN. This network is now referred to as the critic. The loss of this network is approximately equal to the Wasserstein distance, thereby providing a metric of how similar the distribution of the generated data and that of the real data are, thereby providing useful feedback during training. This is in comparison to the original GAN, where loss simply shows the relative strength of each network.

In the WGAN paper, the lipschitz constraint is enforced by clipping the weights of the critic network to ensure the norms of the gradients within the network are at most 1, however this limits the strength of the critic network. To remedy this, Gulrajani et al. \cite{gp} propose using Gradient Penalty, whereby an additional term is added to the critic loss, penalizing the model if the gradients deviate from the target norm value of 1. The gradient penalty term is:

$$ G_{p} = \mathbf{E}_{\hat{x}\sim P_{\hat{x}}}[(\norm{\nabla_{\hat{x}}f(\hat{x})}_{2}-1)^{2}] $$

Where $P_{\hat{x}}$ is the distribution of interpolations formed by sampling genuine and generated data along a normal distribution, and f being the function learned by the critic network. The modified WGAN is referred to as WGAN-GP.

As shown by figure \ref{fig:theta}, with Wasserstein providing a continuous usable gradient everywhere, it does not suffer the vanishing gradient of the GAN as originally formulated by Goodfellow et al. Thus, there is no longer the issue of delicately balancing the strength of generator and discriminator network. Indeed Arjovsky recommends we train the critic network to convergence prior to each generator update. From empirical comparisons \cite{gp}, WGAN-GP is much more stable to train than the original GAN and WGAN with the losses converging much more consistently. 

\subsection{Summary}
This chapter gives a brief introduction to the principles of Convolutional Neural Networks within the context of their application to temporal data. Beyond this, the chapter also gives an outline of the GAN concept, the strengths and weaknesses of the original formulation, and then goes on to describe a later improvement on the concept, the Wasserstein GAN.

\chapter{Introduction to the Dataset and Rendering Motion Results}
\section{Dataset}
The motion capture dataset we used throughout the project was created by Nakamura et al. \cite{nak} in a paper examining transport strategies. The paper examines how object size, weight, and start and end position as well as the need to balance the object affects the use of three possible strategies by trial participants: unimanual transport, handing off between hands, and symmetric bimanual transport. 

We use data from 13 of the participants recorded in the paper. Motion capture data is recorded from 15 optical markers positioned on the participant's body, with 4 around the participant's head, a marker on each shoulder, a 'C7' marker just below the centre of the shoulders, 4 markers around the waist, 1 on each hand, and 1 on each foot. There was also an additional marker on the bowl being transported itself. Positional data is captured from the markers at a rate of 119.88 frames per second, with each marker reporting an x,y,z coordinate in world-space, thus for a trial that is 10 seconds long, we have a data frame of size 1199 frames $\times$48 marker coordinate features.

\begin{figure}[htbp!]
\centering
\includegraphics[scale=0.3]{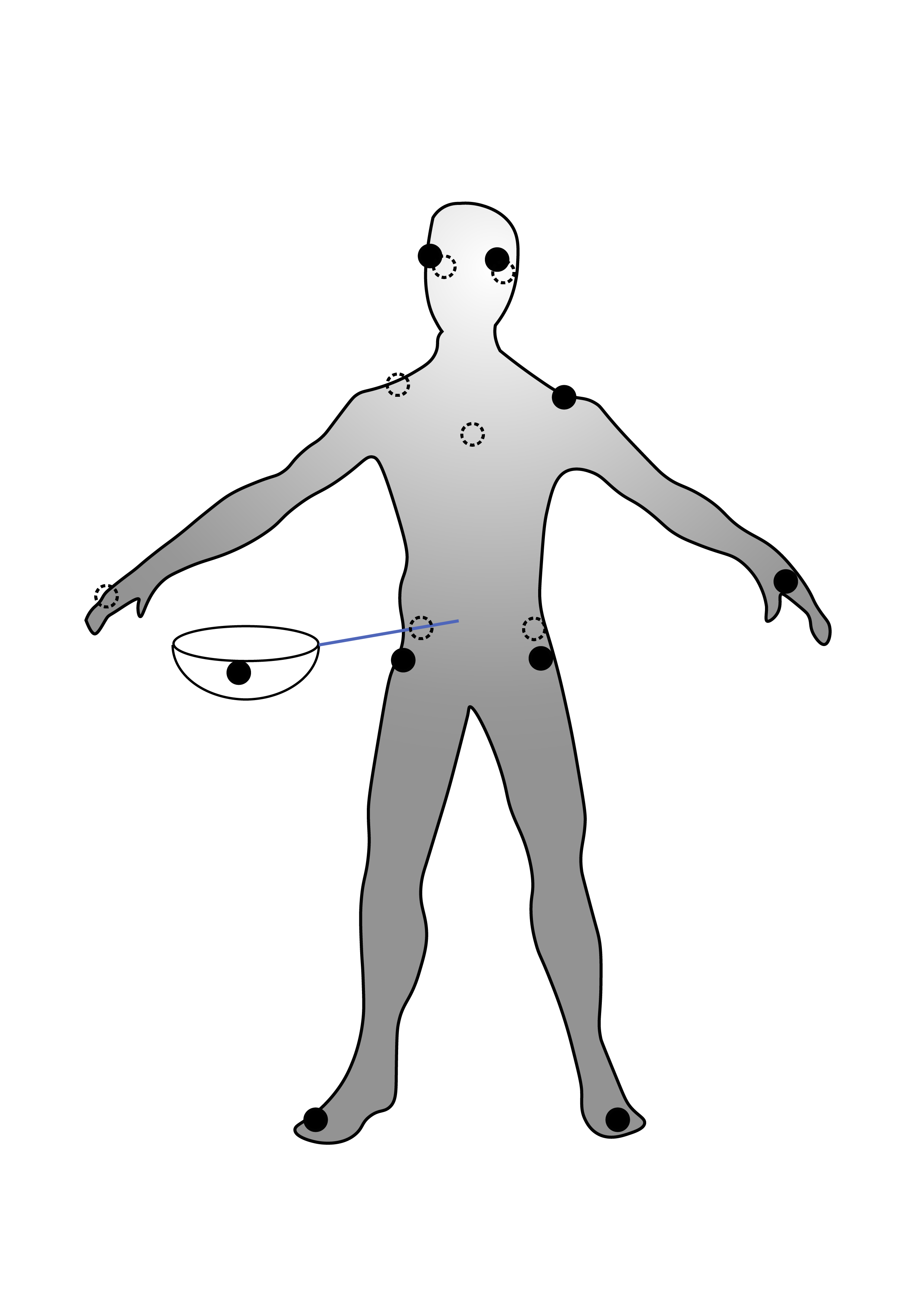}
\caption{positions of motion capture marker features}

\end{figure}
\FloatBarrier

Each of the trials within the dataset consist of a participant moving an IKEA 'blenda' bowl from one table to another. There are 4 size classes of bowl, with the smallest of size 12.2 cm $\times$ 6.1 cm, medium at 20.2 cm $\times$ 9 cm, large at 28 cm $\times$ 13 cm, and largest at 36 cm$\times$ 17.9cm. For each size of bowl, weight is added to make up 3 weight conditions of bowl, with 'heavy' weighing 640 g, 'heavier' weighing 1140 g, and 'heaviest' at 1640 g. For the largest bowl, its empty weight was greater than 640 g, thus there are no trials with the largest bowl in the 'heavy' condition. There are also 2 balance conditions: balanced (simulated by an upright cardboard tube in the bowl); and unbalanced.

Each participant carried out the task under each of the 22 conditions from 3 different starting orientations: directly facing the bowl; facing left of the bowl; and facing right of the bowl. From the 858 trials available to us, we are able to make use of 805 of them for analysis and motion generation, as 53 of the trials are missing data from the 'C7' marker.

A number of different strategies were observed among the participants as they completed the task under each condition. The strategies observed were: A - unimanual left hand; B - unimanual right hand; C - left to right hand-off: D - right to left hand-off; E - left hand to bimanual; F - right hand to bimanual; G - bimanual; H - bimanual to right hand; I - bimanual to left hand. 

The table below shows the frequency at which each strategy was employed, throughout the 805 trials we use for analysis.
Strategies B and G are most dominant throughout the trials. 

\begin{table}[htbp!]
\centering
\begin{tabular}{|c|c|lll}
\cline{1-2}
\multicolumn{1}{|l|}{\textbf{Strategy}} & \multicolumn{1}{l|}{\textbf{Frequency}} &  &  &  \\ \cline{1-2}
A                                       & 40                                      &  &  &  \\ \cline{1-2}
B                                       & 227                                     &  &  &  \\ \cline{1-2}
C                                       & 94                                      &  &  &  \\ \cline{1-2}
D                                       & 44                                      &  &  &  \\ \cline{1-2}
E                                       & 31                                      &  &  &  \\ \cline{1-2}
F                                       & 34                                      &  &  &  \\ \cline{1-2}
G                                       & 318                                     &  &  &  \\ \cline{1-2}
H                                       & 13                                      &  &  &  \\ \cline{1-2}
I                                       & 4                                       &  &  &  \\ \cline{1-2}
\end{tabular}
\caption{Transport Strategies across all trials}
\end{table}
\FloatBarrier
\section{Rendering Motion (for motion synthesis task)}

\begin{figure}[h]
	\centering
	
		\makebox[\textwidth][c]{\includegraphics[width=1.35\textwidth]{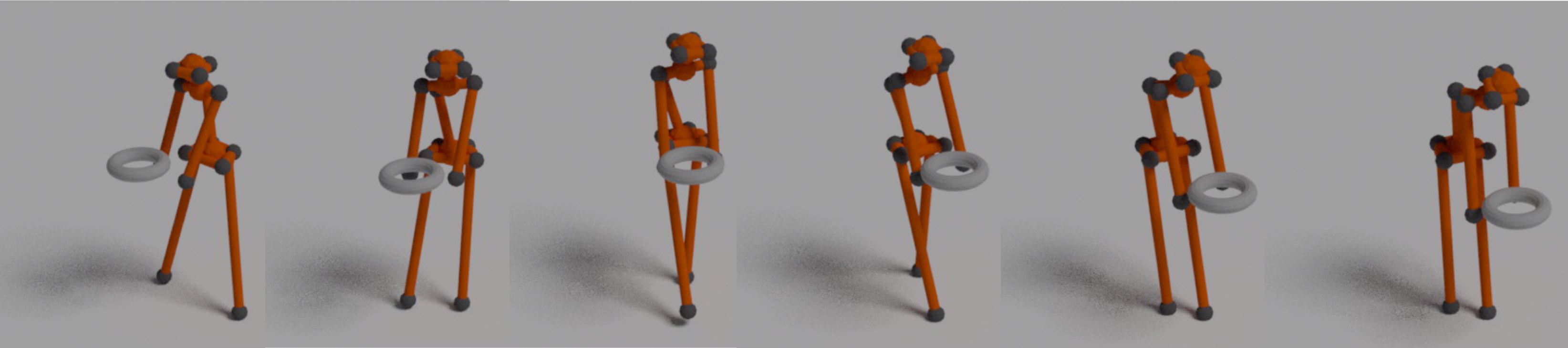}}
		
		\caption{sample motion sequence from real trial data}

\end{figure}

For rendering the model results, we decided to use the open source 3D modelling package Blender, as it readily allows for python scripting. We developed a python script that takes as input a CSV file containing the motion path generated by our trained generative model. 

The script first clears any pre-existing objects from the scene (aside from the camera object and any lighting), after which a grey plane is placed in the centre of the scene. The script then creates a stick figure for the fictional participant, made up of the generated 15 marker positions. Each marker position is visualized as a sphere at the corresponding world-space coordinate. The 15 markers are then joined by fixed length cylinders. Finally a `pelvis' sphere is placed centrally between the 4 hip markers, and a `head sphere' centrally between the 4 head markers.

For each frame of the motion path, the script updates the position of the stick figure. For the cylinders, first the line vector is taken from between the coordinates of the two markers that the cylinder joins. The angle of this vector in world-space is computed, and the cylinder is then positioned at the centre of the vector with the same rotation. 

The bowl marker is represented by a silver torus primitive. The inferred sections of the figure are coloured orange, whilst the generated marker positions are black.

\chapter{Analysis of Motion}

In this section, we outline deep learning techniques that can be successfully used for classifying specific characteristics of an object transport motion sequence.

\section{Classification Tasks}
We establish 3 separate classification tasks to perform on the dataset, 2 binary and 1 multiclass task. Each task is associated with a difficulty corresponding with how easy it is for a human observer, witnessing the motion, to classify the condition.
\newline

The tasks are
\begin{itemize}
    \item Task 1 (binary): determining what weight of object a participant is carrying, between heavy and heaviest condition among bowls of all sizes. This task is considered the hardest to classify by eye.
    
    \item Task 2 (binary): determining if the participant is trying to balance the bowl or not when transporting it. This task is considered easier than Task 1.
    
    \item Task 3 (multiclass): determine what strategy, out of the top 5 most frequently used strategies, the participant is using to carry the bowl. This task is trivial when done by eye. 
\end{itemize}

\section{Data Preprocessing}

Before passing the data to the networks, for each sample, frames before and after the motion are removed, by determining when the bowl is in motion. To standardize the length of each trial, we sample every 12th frame of data (reducing the sampling rate to around 10 fps) from the centre of the shortened data, until we had 32 frames, giving an overall view of around 3.2 seconds of the trial when the bowl is in motion. This data was then normalized by Z-scoring$ z = \frac{x -\mu}{\sigma}$, where $\mu$ is the mean value across all coordinates at all time steps in all samples, and $\sigma$ is the standard deviation. This is done with respect to each individual feature from the marker coordinates, across all samples in the train sets.

\section{Data Augmentation}

Typically, learning tasks that incorporate the use of deep learning need a large dataset to train on. Although there are 805 trials available, for the weight classification task, after filtering for only 'heavy' and 'heaviest' trials, and then negatively down-sampling on the more prevalent 'heaviest' condition, we are left with 436 trials for training and validation. In order to account for this small number of samples compared to the large number of trainable parameters within our neural network, we explore how we can effectively use data augmentation on a motion capture dataset.

We thus create two datasets for training for the weight classification task. The first made up only of the genuine 436 samples, from which we randomly choose 50 samples for validation. In the second dataset, we first sample for our validation set of size 50. The remaining 386 is augmented from the first by a factor of 10 to a size of 3860 samples. We design 3 augmentations: a translation along the X and Y axis of a random distance of ±20 cm; scaling the marker positions by a factor between 0.85 and 1.15 from a centre point between the participant's shoulders and hips; and rotating the motion paths from the bowl starting point randomly by 0-60 degrees. 
Each of these augmentations introduce increased variation among the data, such as by simulating more participants of different sizes, without affecting the core characteristics of the underlying motion.

\begin{figure}[htbp!]
\centering
\includegraphics[scale=0.25]{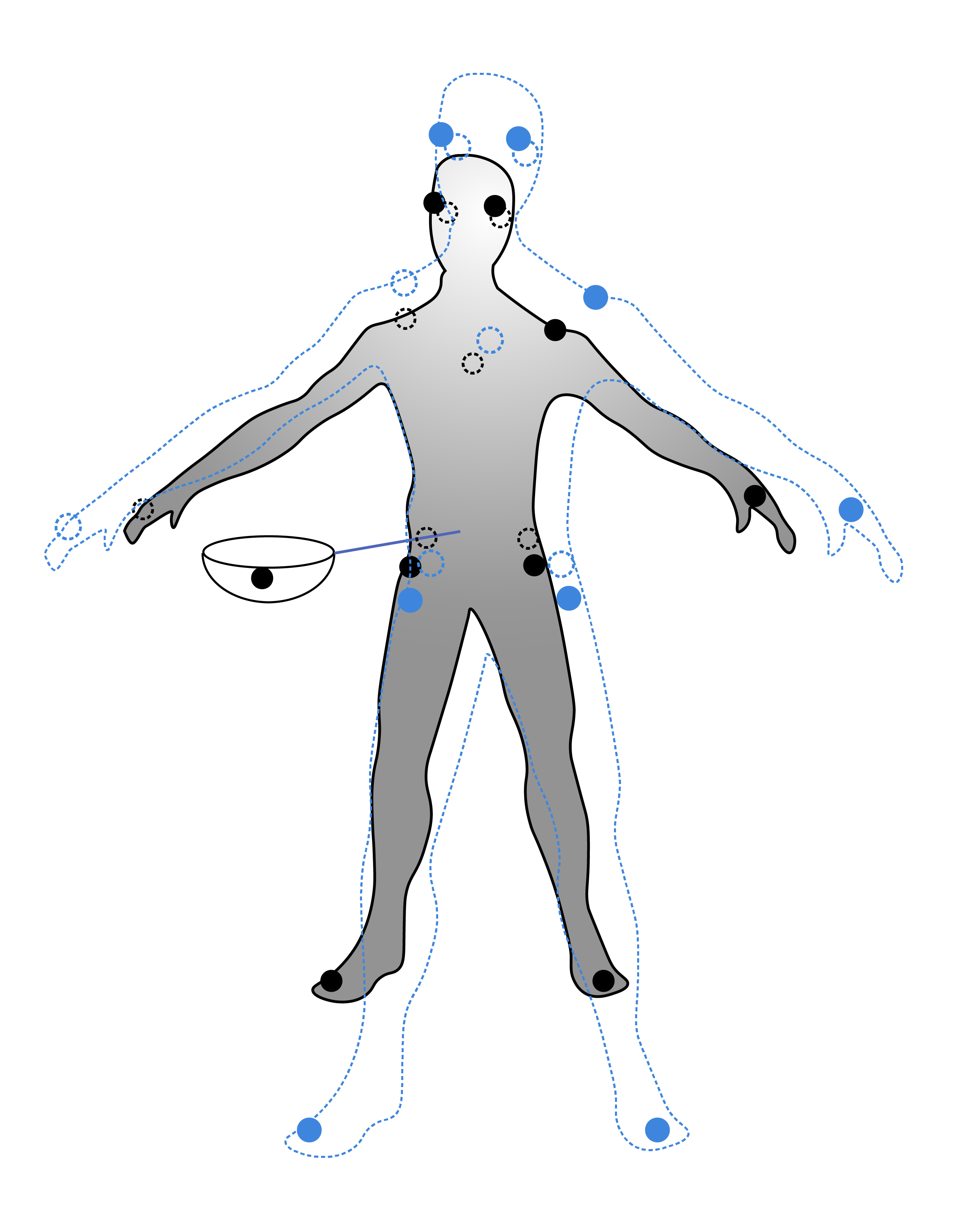}
\caption{Augmentations over a datapoint include scaling marker positions at each time-step from computed central point by a fixed amount}
\end{figure}
\FloatBarrier

In order to trial the efficacy of the augmentations we try to learn task 1 with both the un-augmented dataset and the augmented dataset. 

\section{Design of network}

Given the large number of coordinate features in the input space, and taking inspiration from Du et al.'s hierarchal network \cite{Du}, the network we use divides the set of motion capture markers into 3 related 'clusters'. The upper body cluster, cluster 1, has 7 markers: the 4 markers on the head; the 2 shoulder markers; and the 'C7' marker. The centre body cluster, cluster 2, has 5 markers: the 2 shoulder markers; 'C7' marker; and the 2 hand markers. Finally, the lower body cluster, cluster 3, has 7 markers: the 4 hip markers; the 'C7' marker; and the 2 markers on the feet.  The markers were subdivided in this way with the idea that the interaction between areas of the body that are similarly placed, such as the left and right shoulder, or left and right hand, gives more of a signal as to how someone is moving than parts of the body that are more distant, such as that the head and the foot. 

For each of the markers on the participant's body, the coordinate position of that marker gave 3 features, an X , a Y and a Z coordinate. Thus, for example in cluster 1 with 7 markers, there are 21 positional coordinate features. As stated before, we take a sample of 32 frames of data, thus our input size from cluster 1 is 32 $\times$ 21. 
\newline

\begin{figure}[htbp!]
\centering
\includegraphics[width=1\textwidth]{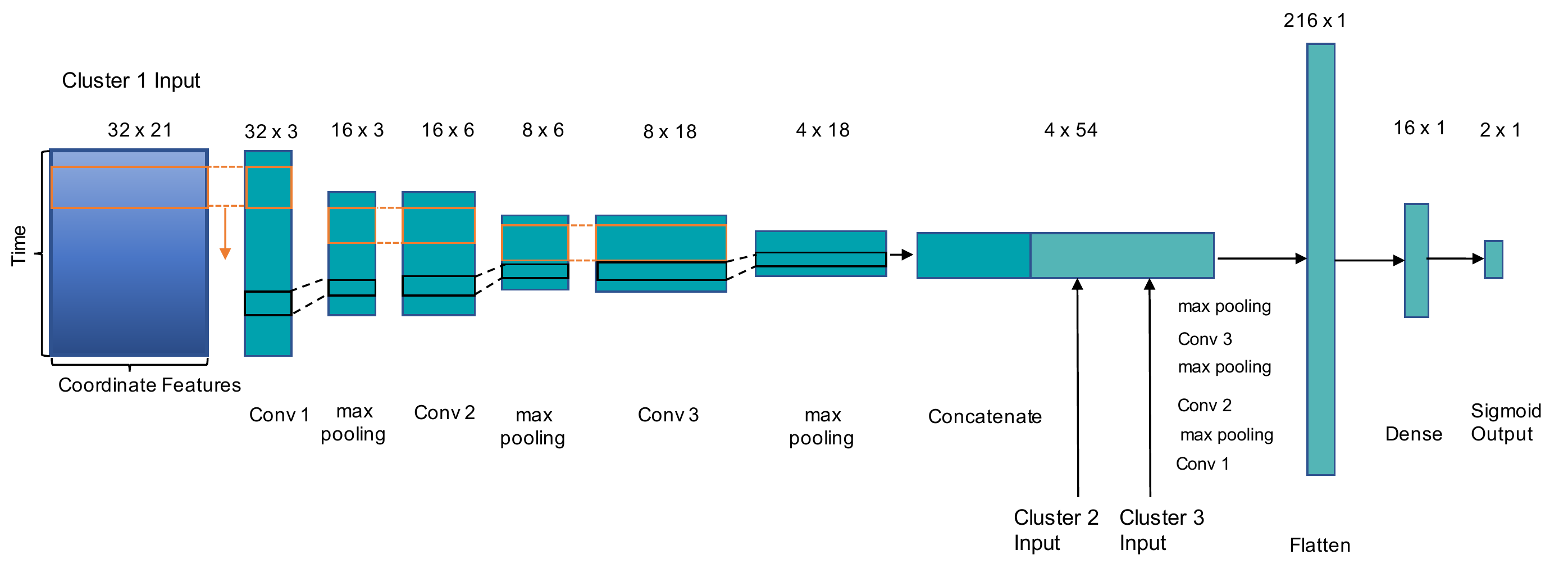}
\caption{Network design focusing on 1 cluster of markers}
\end{figure}
\FloatBarrier
For each cluster, we have an input tensor of size 32 $\times$ k where k is the number of markers in that cluster multiplied by 3. We pass the 3 inputs through a series of separate one dimensional convolutions and max pooling operations, before concatenating the 3 outputs. We then run this concatenated tensor through a dense layer, before passing it to the final output layer.

After each convolutional layer, a max-pooling operation reduces the number of time-steps in the output by half. With a pool size of 2 time-steps, the pooling layer outputs the largest value from every pair of subsequent time steps. Pooling is used to reduce the overall number of parameters within the model, and reduce the computational complexity of training.
In addition to this, we make use of dilated convolutions in the first layers of the network, with a spacing size of 1. 

\begin{figure}[htbp!]
\centering
\includegraphics[width=0.6\textwidth]{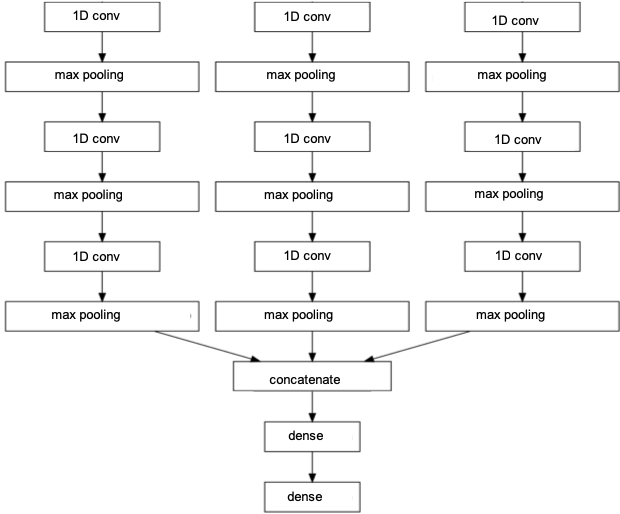}
\caption{The network structure for analysis tasks.}
\end{figure}
\FloatBarrier

This network, and the networks used for motion synthesis, are implemented using Tensorflow and Keras. Tensorflow is a deep learning framework, while Keras is a high level API that sits on top of Tensorflow.

\section{Classification Tasks - Results}

Using the network described above with ReLU activation functions in all convolutional layers, and training using the ADAM optimizer function across 400 epochs of data, these are the results we obtain across the 3 tasks. 

\subsection{Weight Classification}

For the un-augmented dataset, as stated before there are a total of 436 samples, from which 50 are taken to be used as the validation set. Both the training and validation set are balanced among the two conditions of 'heavy' and 'heaviest'.

Looking at the validation loss it is apparent that the network over-fits on the training samples, as the validation loss quickly diverges from training loss and increases.

\begin{figure}[htbp!]
\centering
\includegraphics[width=0.5\textwidth]{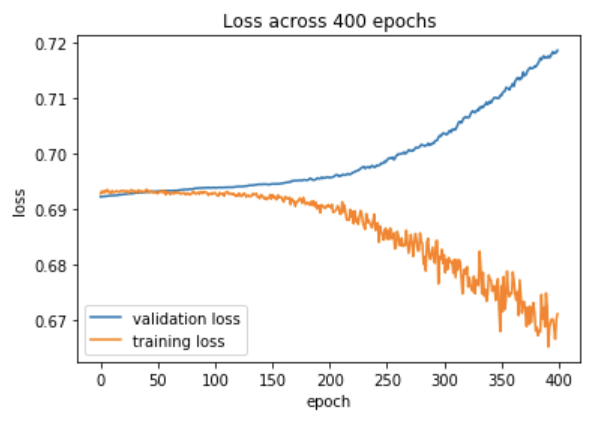}
\caption{Loss using un-augmented dataset on weight classification task}
\end{figure}
\FloatBarrier
This is somewhat unsurprising. As the network has 5,210 trainable parameters, the network simply 'remembers' the 386 training samples, resulting in a poor performance on the unseen validation set, with the accuracy in the classification task remaining at chance level.

Increasing the dropout rate of the network, essentially turning off a percentage of the neurons, randomly, within the network during each batch of training is not enough to counteract this lack of data.

Turning to the implemented augmentations over the dataset, we first examine how the network performs with 1 augmentation, rotations. Boosting the dataset by a factor of 4, we can see evidence of the regularizing effect of the augmented data. Although loss on the validation set initially increases and peaks at epoch 150, after this there is a steady decrease, following the trend of the training loss. After 400 epochs, accuracy is slightly above chance at 54\%.

\begin{figure}[htbp!]
\centering
\includegraphics[width=0.5\textwidth]{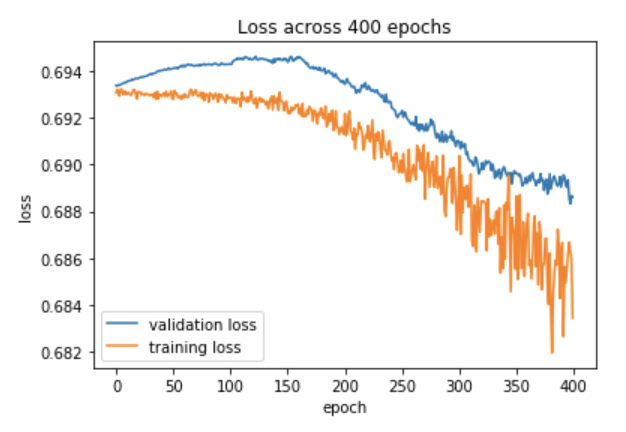}
\caption{Loss using dataset with 1 augmentation (rotation) on weight classification task}
\end{figure}
\FloatBarrier
Finally, using the dataset with all 3 augmentations, and a total of 3860 training samples, we are able to achieve a constant decrease in loss that corresponds with a much improved accuracy over the classification task, peaking at epoch 150, with 70\% correct classifications on the validation set. 

\hspace{-0.5em} \begin{figure}[h]
	\centering
	\begin{minipage}[t]{5cm}
		\centering
		\includegraphics[scale=0.5]{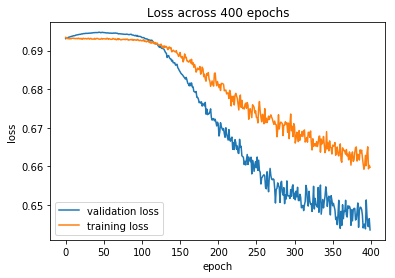}
		\caption{loss using fully augmented dataset on weight classification}
	\end{minipage}
	\hspace{3cm}
	\begin{minipage}[t]{6cm}
		\centering
		\includegraphics[scale=0.5]{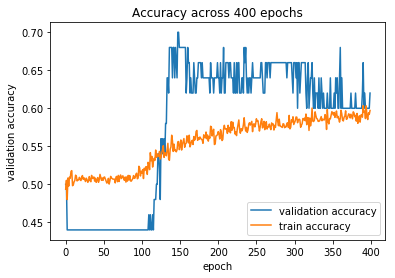}
		\caption{accuracy using fully augmented dataset on weight classification}
	\end{minipage}
\end{figure}
\FloatBarrier
Given the successful result using augmented data, we augment the training data used for subsequent classification tasks using the aforementioned techniques.

\begin{figure}[htbp!]
\centering
\includegraphics[width=0.5\textwidth]{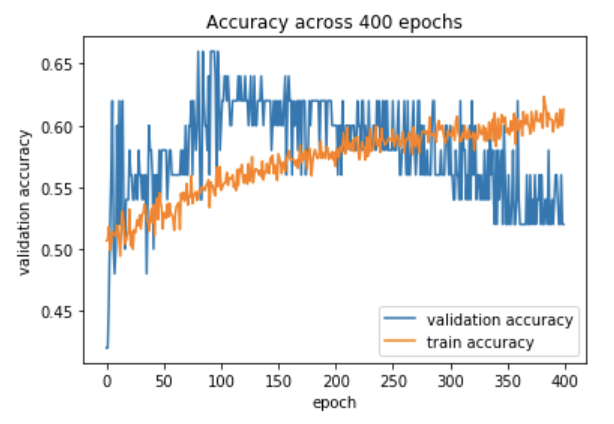}
\caption{accuracy using LSTM model with fully augmented dataset on weight classification}
\end{figure}
\FloatBarrier
We compare the result we achieve with a convolutional model to that of a single cell LSTM, a standard model typically used for tasks involving the analysis of temporal data. Across 400 epochs, the LSTM model accuracy peaks at 66\%, before steadily decreasing.

\subsection{Balance Classifier}

For this task, the dataset is already balanced, thus we have 805 samples, from which we use 100 samples as the validation set, and augment the rest to a size of 7050 samples, with which we train the network. With the regularizing effect of the augmented data, in addition with dropout, loss on the validation set decreases steadily over the 400 epochs, with task prediction accuracy peaking at 87\%.

\hspace{+0em}	 \begin{figure}[htbp!]
	\centering
	\begin{minipage}[t]{5cm}
		\centering
	\includegraphics[scale=0.5]{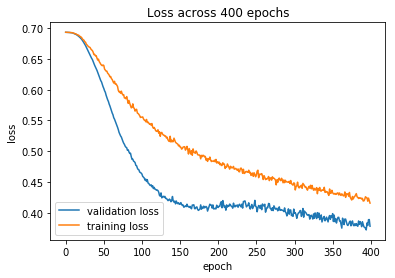}
	\caption{\newline loss on balance classification}
	\end{minipage}
	\hspace{3cm}
	\begin{minipage}[t]{6cm}
		\centering
		\includegraphics[scale=0.5]{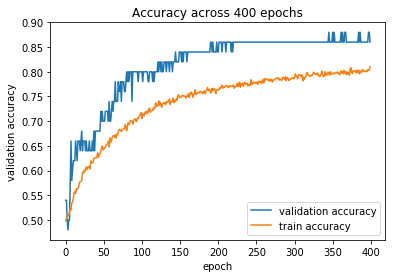}
		\caption{accuracy on balance classification}
	\end{minipage}
\end{figure}
\FloatBarrier

\subsection{Strategy Classifier}

 As previously mentioned, the frequency in which strategies are employed varies drastically. For the strategy classifier, we choose the top 5 most frequent strategies A, B, C, D, G for the learning, so as to ensure the network has enough samples to train on. 
 Our remaining dataset has 723 samples, from which we take 100 samples for the validation set, and again augment the rest by a factor of 10 for use as training.
 
Running for 400 epochs, loss decreases steadily on both training and validation data.

\begin{figure}[htbp!]
\centering
\includegraphics[width=0.5\textwidth]{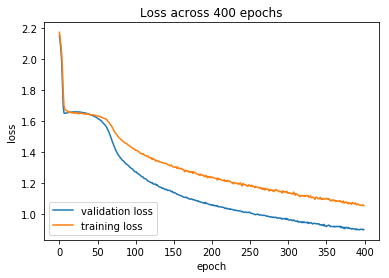}
\caption{Loss using dataset (with augmentation) on strategy classification}
\end{figure}
\FloatBarrier
Overall the samples within the validation set, we achieve an accuracy of 78\%. The prediction accuracy for each particular strategy class is further broken down below in a confusion matrix.

\begin{figure}[htbp!]
\centering
\includegraphics[width=0.6\textwidth]{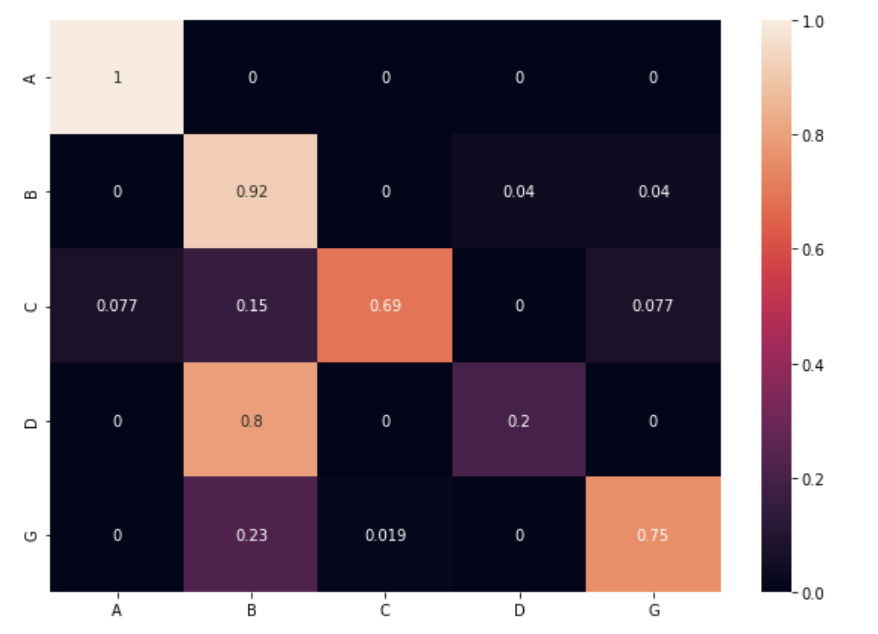}
\caption{Accuracy for strategy classification task (rows are actual labels, columns are predicted labels), with values being proportion of a label being predicted per class}
\end{figure}
\FloatBarrier
As shown in the confusion matrix, the network successfully learns to classify strategies A (unimanual left hand), B (unimanual right hand), and G (bimanual) with an accuracy of 100\%, 92\% and 75\% in classifying each respectively.  

For less prevalent strategies however the results are more mixed.
For strategy D, the model achieves only 20\% accuracy, with the model confusing the remaining 80\% to be strategy B. Interesting to note is the model's higher success rate in classifying strategy A compared to D, even though it occurs less frequently, appearing 40 times, compared to 44 for D. The two highest accuracy values within the matrix correspond with the two unimanual strategies, A and B.

\subsection{Summary}

In this chapter, we propose a hierarchical style convolutional network, that can be used to successfully classify not only how an individual is performing an object transport motion, but also classify characteristics of the object they are transporting.
The chapter also highlights the efficacy of our implemented data augmentation techniques, showing how such augmentations are vital when learning with a small motion capture dataset.

\chapter{Motion Synthesis}

For the motion synthesis task, data preprocessing is done in much the same way as in the classification tasks, with the only difference being how we sample frames within each sequence. In order to capture the entire movement within a trial, rather than take every 12th frame from the centre of the motion to get 32 frames, 32 frames are sampled evenly from across the entire duration of the movement. This allows the entire movement to be summarized within the 32 frames. However information related to the speed of the motion is lost. Rendering this data with a fixed frame-rate tends to lead to `pinching', whereby small movements at the start of the motion are relatively slow, while large movements in the centre of the motion are relatively fast. 

Besides this preprocessing, the data augmentation performed on the training set is more aggressive, given the tendency of large networks, such as DCGANs to overfit. Thus, 795 genuine trials are augmented by a factor of 27 to give 21465 training examples, using the same augmentations described for the classification tasks.

\section{DCGAN}

The generator network for the DCGAN takes in a vector of random noise with a mean of 0, standard deviation of 1, and size of 100. This random noise is first projected onto 1536 nodes of a dense layer, to give it an equal number of features as the genuine input data (32 timesteps $\times$ 48 marker coordinates). The 1536 output of the layer is then reshaped to (4$\times$384). This is then passed through a 1-D up-sampling operation with an up-sampling factor of 2. This duplicates each of the time steps, to take the input shape to (8x384). This is then passed through a 1D convolutional layer with 192 filters to transform the data shape to 8x192. This sequence of up-sampling and 1D convolutions is repeated twice more, with each convolutional layer having half as many filters as the previous layer. The final output shape is 32$\times$48, equal to that of the true data (32 recorded time-steps, with 48 x,y,z coordinates for 16 markers).

\hspace*{-0.1cm}\begin{figure}[!htbp]
	\centering

		\makebox[\textwidth][c]{\includegraphics[width=1.3\textwidth]{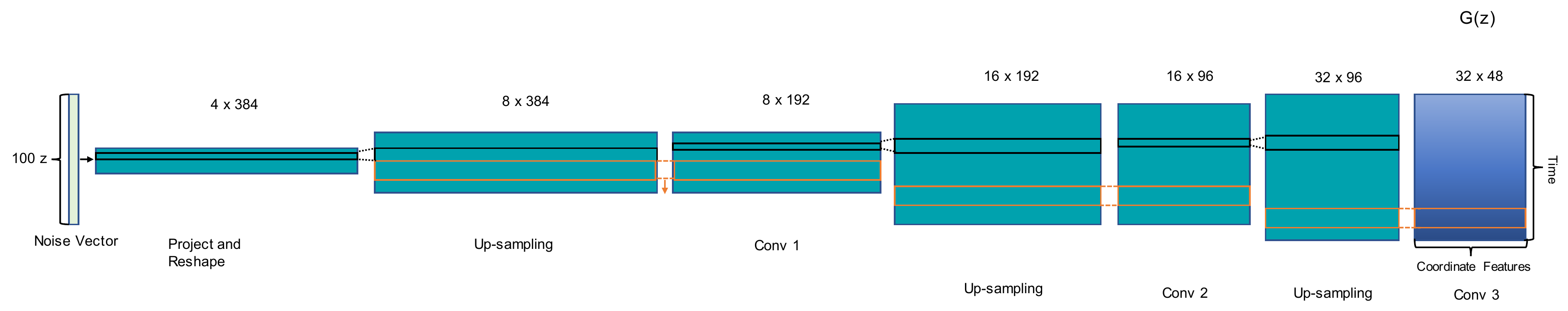}}
		\caption{Structure of the generator network. z is the noise input, and G(z) is the generator function}

\end{figure}
\FloatBarrier

The discriminator network we use is similar in structure to the network used within the motion recognition tasks. However, as opposed to passing the motion capture marker data in as multiple clusters, all of the marker data is passed into the discriminator network as a single combined input of 16 markers, or 48 total x,y,z coordinates. The structure of the discriminator mirrors that of the generator, with 3 convolutional layers, the first having 96 filters, the second having 192, and the third having 384 filters. During the training of the DCGAN, the input to the discriminator was a batch of size 64 of alternately real motion data, or data created by the generator network. Within each convolution, a stride of size 2 was used to halve the size of the output at each convolutional layer and thus reduce the total number of parameters and thereby reduce training time. 

The hyper-parameters of each network were initially set to those recommended by the original DCGAN paper \cite{alec}. In accordance with the paper, batch normalization was also used after each convolutional layer in both networks to increase the speed of training. 

A well known issue when training a GAN is overconfidence and that the discriminator becomes reliant on a small set of features to detect real images. Accordingly the generator may simply produce these corresponding features to exploit this, ultimately harming the overall training.
Consequently, we penalize any output made by the discriminator that goes above 0.9 with regard to predicting a genuine sample, as recommended by Salimans et. al \cite{sal}, by setting the target labels for genuine data in the discriminator as 0.9.  

\subsection{DCGAN results}

With this DCGAN configuration, it is obvious from the loss curve below that the discriminator quickly overpowers the generator. As the discriminator loss falls, the generator loses signal as to how to correct its output, thus the loss for both discriminator and generator become stationary after approximately 40 epochs. 

\begin{figure}[!htbp]
	\centering

		\includegraphics[scale=0.6]{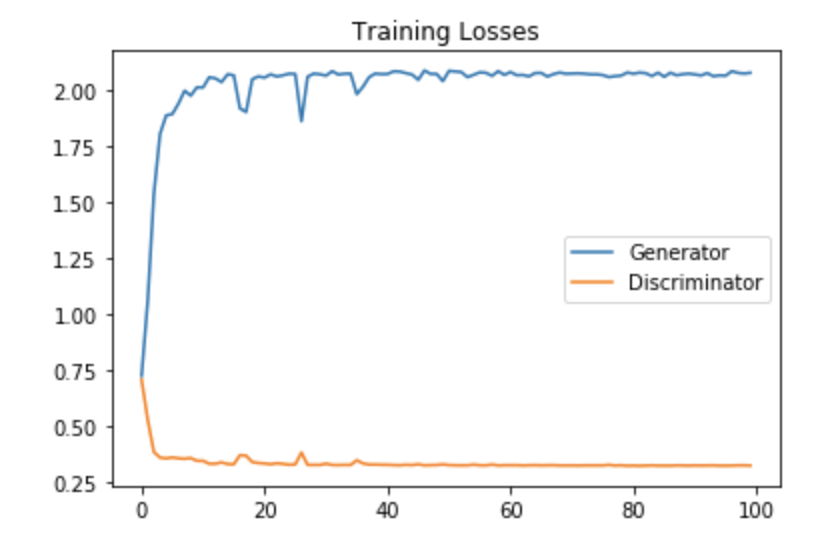}
		\caption{Loss across both networks for DCGAN model}

\end{figure}
\FloatBarrier

At epoch 20, the generator is able to produce a reasonable estimation of motion. The network learns the trajectory of the bowl, whilst the stick figure rendered is mostly anatomically correct. However, further training the network does not yield better results, indeed at epoch 50, the figure generated barely resembles a human. Interestingly, as shown above, there is little change in the discriminator and generator loss between these epochs, highlighting how the GAN loss is a poor metric during training, and that the only way to see how the network is performing is via manual inspection. 

\begin{figure}[!htbp]
	\centering

		\makebox[\textwidth][c]{\includegraphics[width=1.35\textwidth]{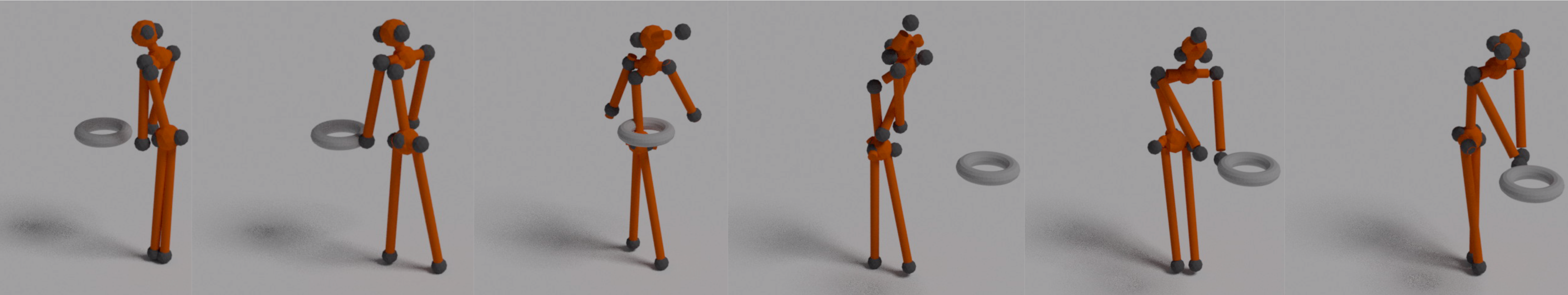}}
		\caption{Motion sequence generated by DCGAN after 20 epochs}

\end{figure}
\FloatBarrier

\hspace{-0.5em} \begin{figure}[h]
	\centering
	\begin{minipage}[t]{5cm}
		\centering
		\includegraphics[scale=0.5]{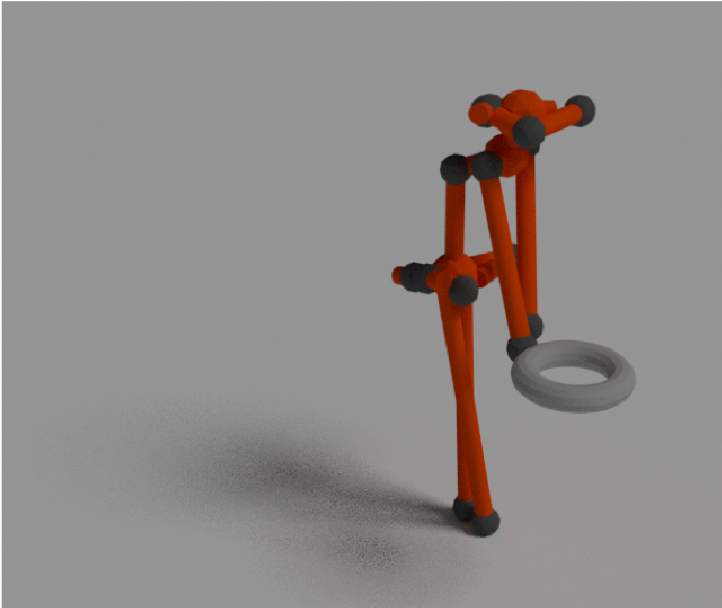}
		\caption{End of sequence after 20 epochs training DCGAN}
	\end{minipage}
	\hspace{3cm}
	\begin{minipage}[t]{6cm}
		\centering
		\includegraphics[scale=0.5]{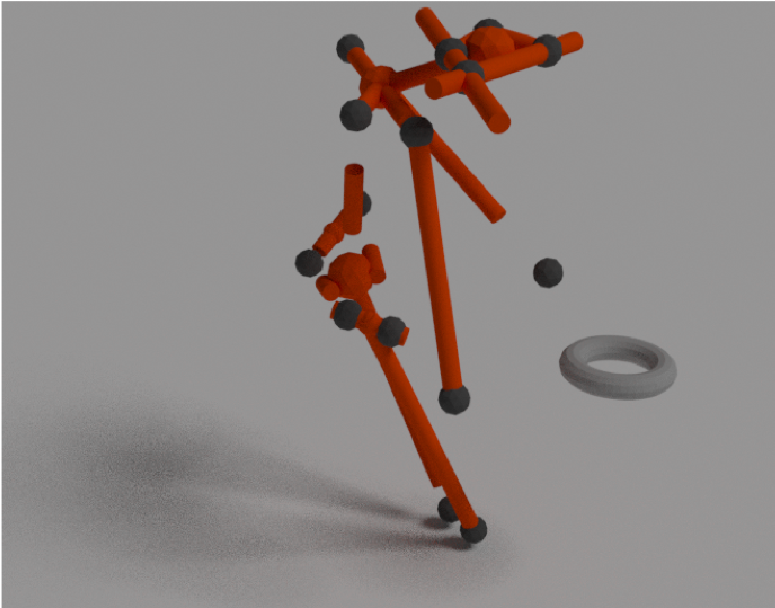}
		\caption{End of sequence after 50 epochs training DCGAN}
	\end{minipage}
\end{figure}

Inferring that the generator loses signal between these 2 epochs, we try and further balance the strength of the 2 networks by increasing the dropout proportion across all layers in the discriminator. However, this did little to remedy the imbalance. Similarly, reducing the number of filters in the discriminator network simply delayed the time taken for the discriminator to overpower the generator, as shown below, whilst the results it produces are similar to the original network.

\begin{figure}[!htbp]
	\centering

		\includegraphics[scale=0.6]{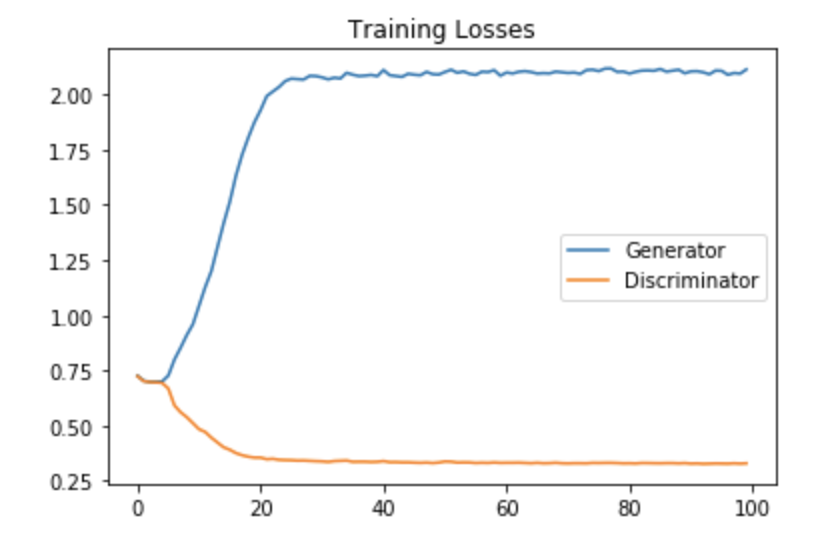}
		\caption{Loss in DCGAN model with high dropout and reduced filters in discriminator}

\end{figure}
\FloatBarrier

These results are somewhat unsurprising. It is notoriously difficult to balance the two networks. If we reduce the discriminator strength, then the generator network has little impetus to produce a reasonable output. As in our case, if the discriminator becomes too powerful however, all learning stops, and the output deteriorates. 

\section{WGAN-GP}

Following this failure of the DCGAN, we trial a WGAN-GP model to overcome the issue of the generator losing signal. The structure of the model is similar to that of the trialled DCGAN model, with the generator having the same layers, and the critic of the WGAN-GP having the same number of convolutions as the critic in the DCGAN. The only difference in network structure is that we no longer use batch normalization in the critic network, as this prevents the approximation of the Wasserstein distance. We first try a network without batch normalization in the generator, and then a network with normalization.

The input for the generator function is a vector of random noise following a normal distribution, with a mean of 0, and variance of 1. As recommended by Arjovsky \cite{arj}, the critic network is trained to convergence prior to each generator update. Thus, for each generator update, we train our critic network on 15 batches of both synthetic and real data. From each set of samples we generate randomly weighted averages between genuine and generated data to create the gradient penalty term.

\subsection{WGAN-GP without batch normalization in generator}

The Wasserstein distance quickly decreases as the model is trained over 200 epochs, converging around epoch 150. Our results are in line with Arjovsky's claims that reduced Wasserstein distance can be directly correlated with improved output quality. As training goes on, the output motion sequences visibly improve in quality. 

\begin{figure}[!htbp]
	\centering

		\includegraphics[scale=0.5]{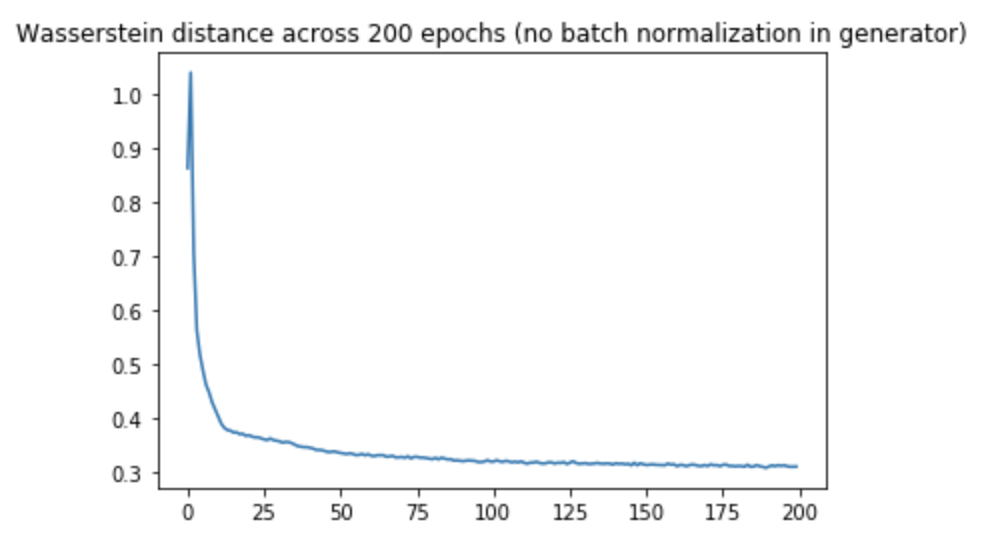}
		\caption{WGAN loss with gradient penalty and no batch normalization in generator}

\end{figure}
\FloatBarrier

\begin{figure}[!htbp]
	\centering

		\makebox[\textwidth][c]{\includegraphics[width=1.35\textwidth]{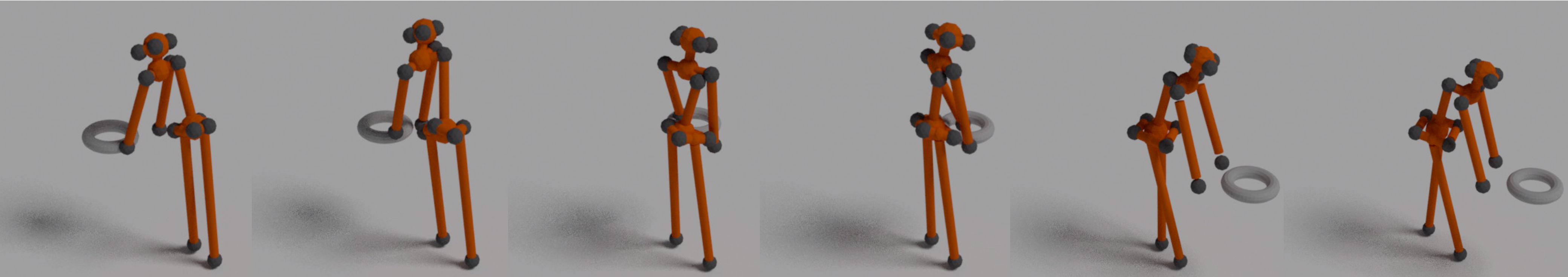}}
		\caption{Bimanual motion, pivoting on right foot, generated by WGAN-GP}
		\label{fig:good1}

\end{figure}

\begin{figure}[!htbp]
	\centering

		\makebox[\textwidth][c]{\includegraphics[width=1.35\textwidth]{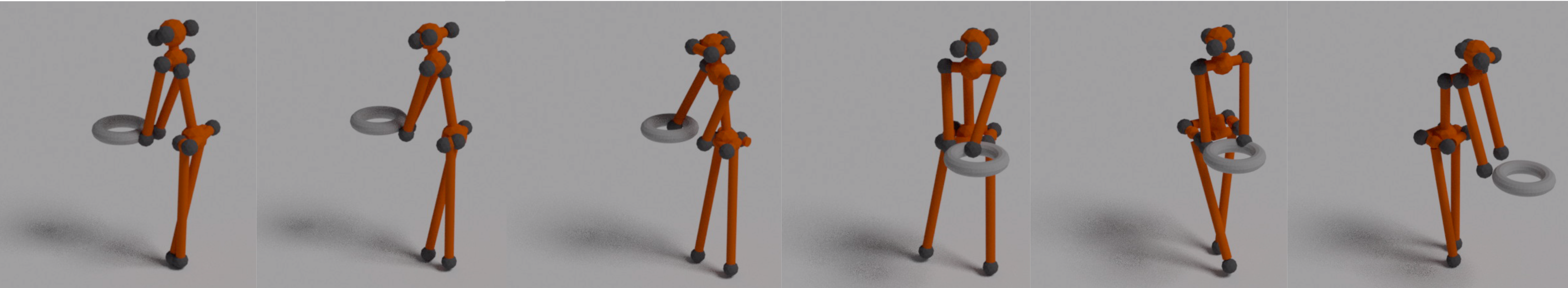}}
		\caption{Mixture of left hand and bimanual strategy, pivoting on left foot, generated by WGAN-GP}
		\label{fig:good2}

\end{figure}

As shown by the results below, the WGAN-GP is successful in lifelike motion sequences of object transportation. Of particular note is that this model did not suffer from 'mode-collapse', a common pitfall when training GANs, where the generator simply outputs one good representation repeatedly. Instead, depending on the noise input, the generator is able to produce multiple strategies, as evidenced by fig. \ref{fig:good1} and \ref{fig:good2} showing different styles and strategy of transportation motion. More examples of different styles and strategies of motion generated by the network are shown in the appendix.

\FloatBarrier
\FloatBarrier

Given that the loss function does not account for the limits of the human body, there are rare instances of impossible motion generated by the network. As shown below, when two incompatible strategies are combined by the network, such as passing the object over the left side of the body, and passing it over the right, this leads to poor results.

\begin{figure}[!htbp]
	\centering

		\makebox[\textwidth][c]{\includegraphics[width=1.35\textwidth]{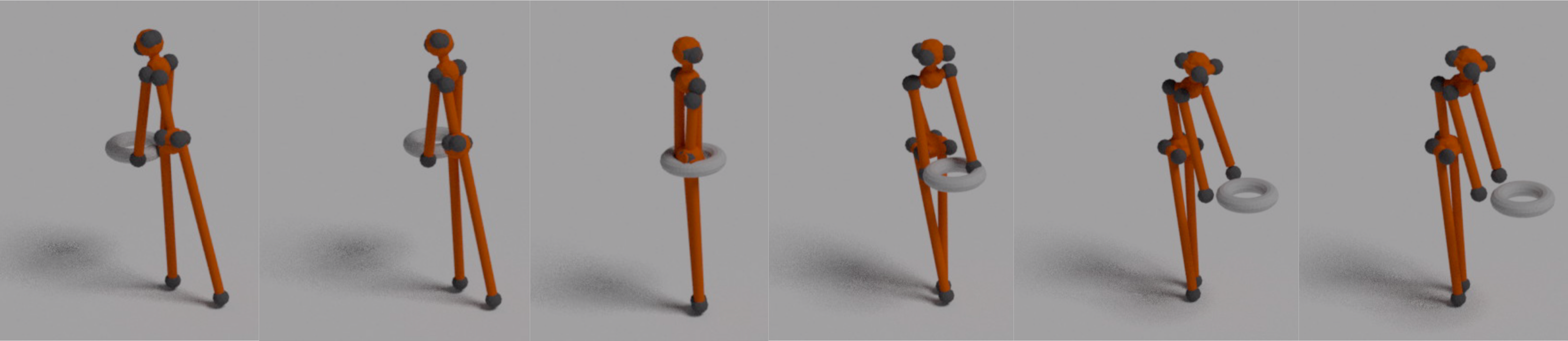}}
		\caption{Impossible motion generated by WGAN-GP}

\end{figure}
\FloatBarrier
\subsection{WGAN-GP with batch normalization in generator}

Our first WGAN-GP model removes batch normalization from both the discriminator and generator networks. we next try a WGAN-GP model with batch normalization in the generator network. However this leads to the Wasserstein distance stabilizing at a much greater value of 1.3 in contrast to the 0.3 achieved by the previous model. 

\begin{figure}[!htbp]
	\centering

		\includegraphics[scale=0.6]{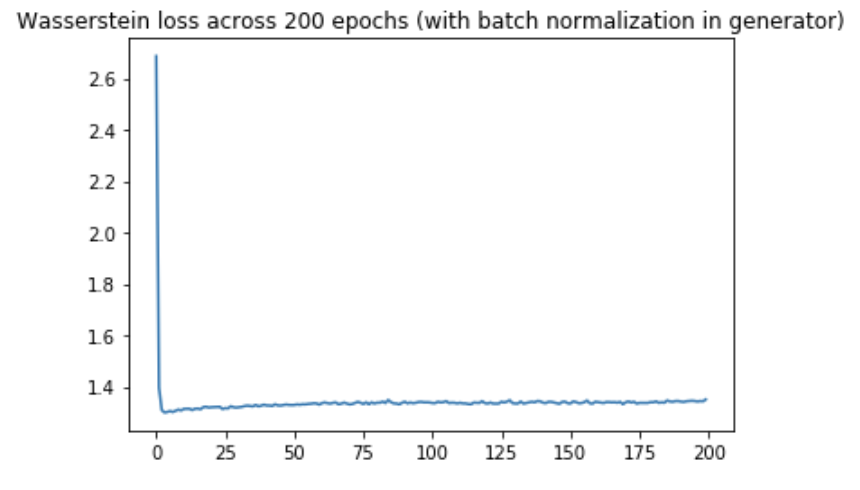}
		\caption{WGAN-GP with batch normalization in generator}

\end{figure}
\FloatBarrier

The motions generated with the network incorporating batch normalization are visually less lifelike than the network without. Marker coordinates appear much more compressed than the previous results, with respect to the X and Y axis. Furthermore, the network undergoes 'mode collapse' generating the same impossible motion for every noise input, wherein the figure passes the bowl through themselves.

\begin{figure}[!htbp]
	\centering

		\makebox[\textwidth][c]{\includegraphics[width=1.35\textwidth]{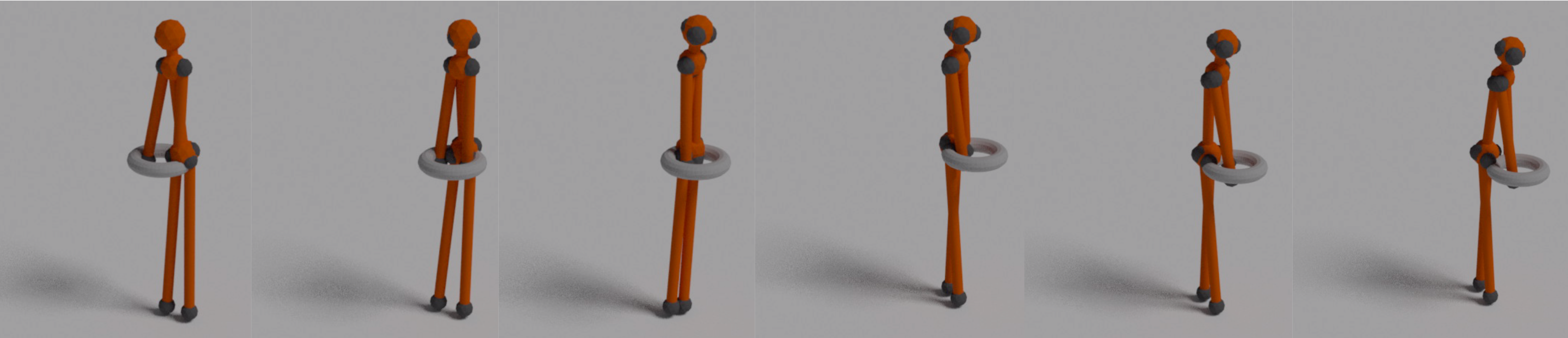}}
		\caption{two-handed motion generated by model with batch normalization in generator}

\end{figure}
\FloatBarrier

\section{Conditional WGAN-GP}

Following the success of our WGAN-GP model, we explore how the model can be modified to represent motions with a certain quality such as balanced motions, or those wherein the participant is carrying a large or small object. 
Following the structure proposed by Mirza and Osindero \cite{mir}, labels y are passed to the generator network with the latent input z, and to the critic network, either concatenated with the genuine motion sequence or concatenated to the generated sample conditioned on the labels y.

\begin{figure}[htbp!]
\centering
\includegraphics[width=0.6\textwidth]{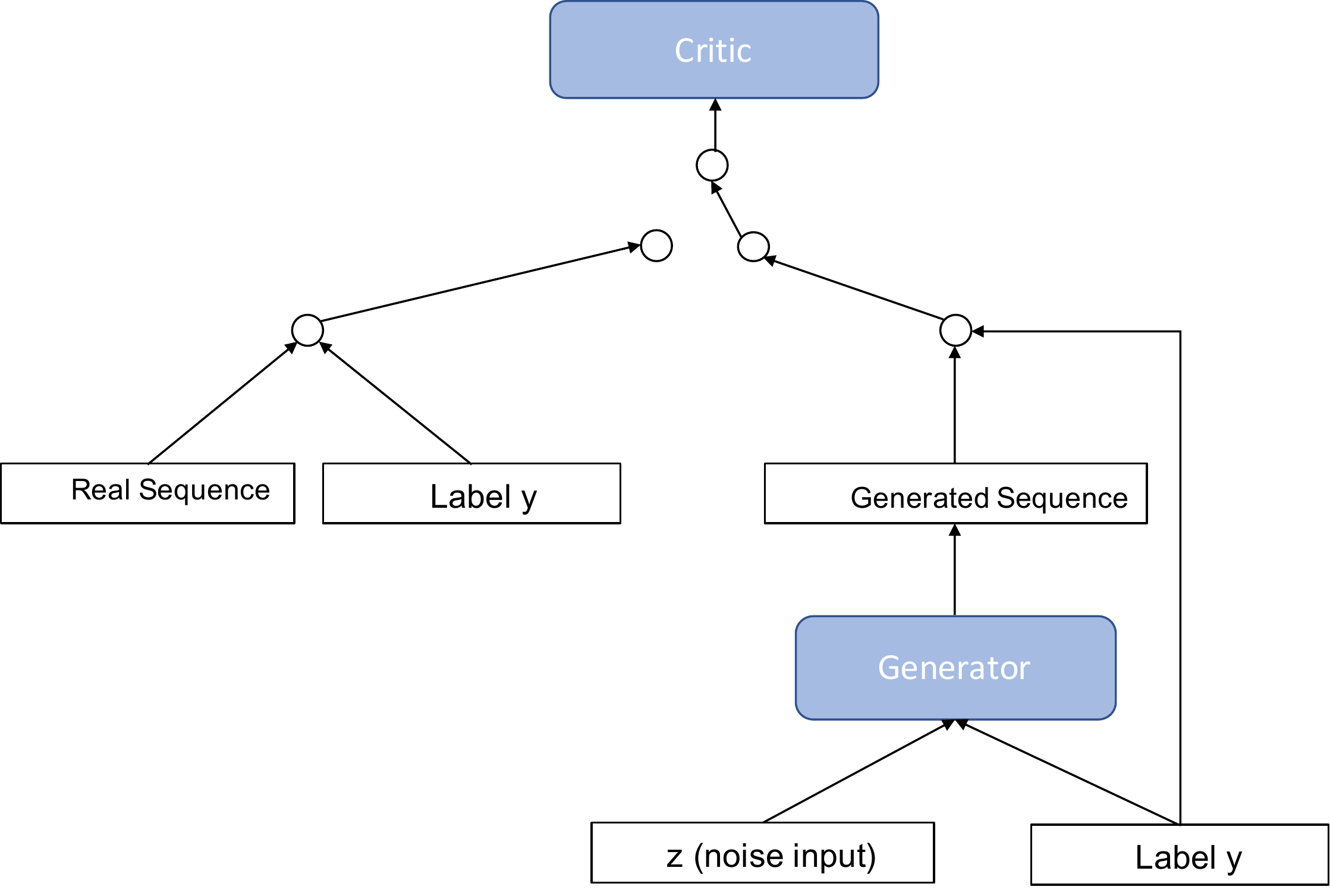}
\caption{How labels are passed in conditional model}
\end{figure}
\FloatBarrier
For this conditional model, 6 classes are defined, made up of the 3 weight classes, as well as the balance parameter, balanced and unbalanced. After training for 120 epochs, and inspecting the output, it is observed that the movement characteristics shown by samples conditioned on unbalanced and light weight bowls is typically different from weighted balanced samples. For a light unbalanced object, movement tends to use a uni-manual strategy. 

\subsection{Conditional WGAN-GP results}

\begin{figure}[!htbp]
	\centering

		\makebox[\textwidth][c]{\includegraphics[width=1.35\textwidth]{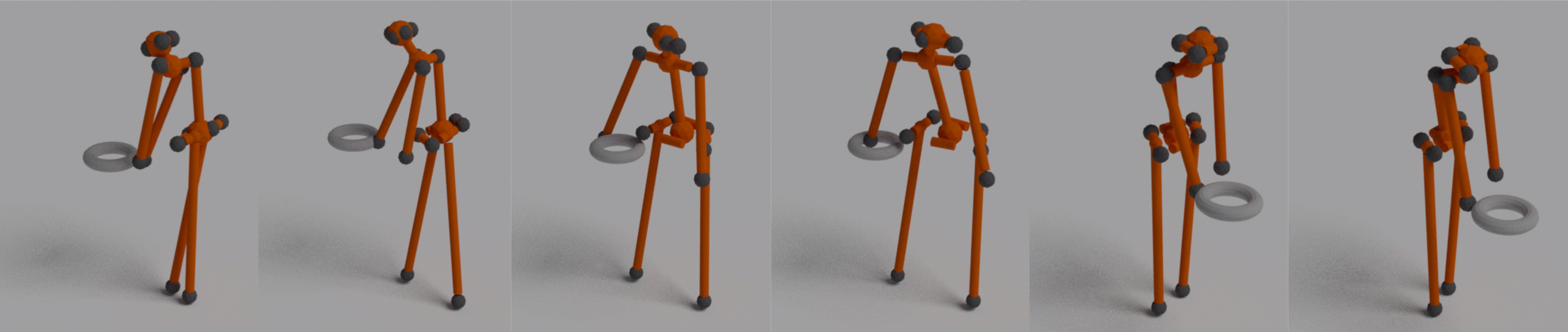}}
		\caption{Motion conditioned on unbalanced and light bowl}

\end{figure}
\FloatBarrier

For a heavy balanced object, the generated samples tend to show more careful bimanual transport strategies, while the figure tends to be more stooped. Unfortunately, as mentioned previously, the frame sampling technique used mean that passing in the label `balanced' to the model does not necessarily generate a motion with the cautious slowness associated with transporting a balanced object.

\begin{figure}[!htbp]
	\centering

		\makebox[\textwidth][c]{\includegraphics[width=1.35\textwidth]{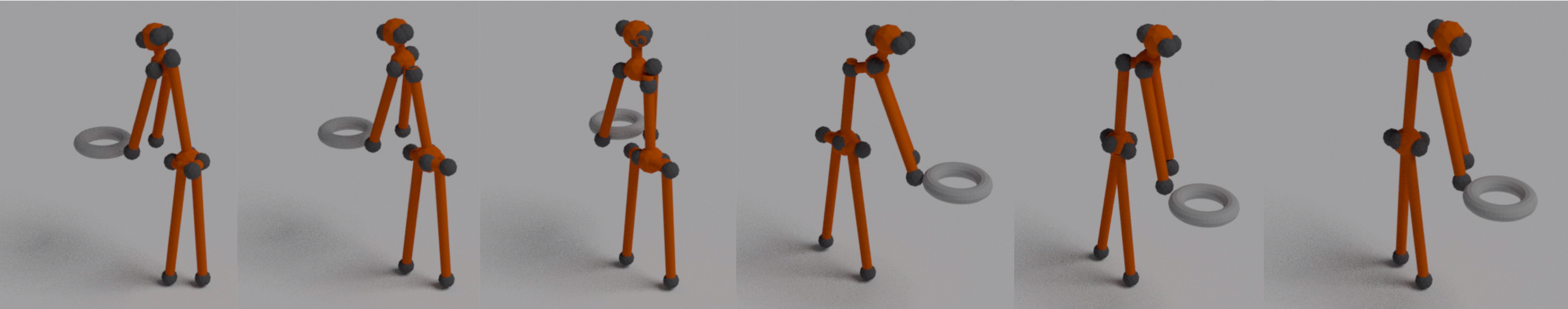}}
		\caption{Motion conditioned on balanced and heavy bowl}

\end{figure}
\FloatBarrier

Nonetheless, in terms of relative marker positions, the conditional model appears to successfully learn stereotypical representations of movement under the constraints of the conditioning parameter.

\subsection{Summary}
This chapter first highlights the instability of training a DCGAN. The chapter then goes on to show  how our proposed WGAN-GP model can be used to successfully generate lifelike motion sequences, while we also show how the model can be conditioned on to output motions with certain characteristics. Besides the success of the proposed model, we also highlight some limitations with the approach we take, such as loss of motion speed information during data preprocessing.

\chapter{Conclusions}

\section{Achievements}

This project had two main ambitions: perform motion classification and motion synthesis. With regard to the first task, there is no current benchmark of performance on this dataset, as it is still relatively unexplored, therefore it is hard to judge the performance of the model proposed. 
Nonetheless, the project does show that small nuances in object transport motion, caused by factors such as if the object being carried is heavy or light, can be learned using a convolutional network. The report also outlines the design of several data augmentation techniques for the application on motion capture sequences, allowing for the use of deep learning on a relatively small dataset.

With regard to the motion synthesis task, although it is difficult to describe the success of a generative model in a quantitative manner, on inspection, the WGAN-GP model we propose successfully produces, for the most part, realistic believable motion sequences. This proposed architecture is able to model the multidimensional space of object transport motion, allowing the synthesis of transport motion sequences of varying styles and strategies, given some noise input. Finally, we are able to show that when this space is parameterized, and the generator network is fed latent noise in addition to these parameters, the motion samples produced seem typical of the movement expected of an individual subject to such constraints. The strengths and weaknesses of particular varieties of GAN are also highlighted within this report.  

The success of both tasks further highlights the efficacy of convolutional models when it comes to learning from motion sequence data.

\section{Future Work}

With regard to the rendering of results, given that the original dataset does not include markers denoting the position of knee and elbow joints, the choice to render the stick figure with rigid cylinders between markers sometimes leads to odd looking results. In order to remedy this, we can explore the use of inverse kinematics, to infer the position of the missing joints, so that the quality of results can be judged more accurately.

Although the number of impossible motions generated by the WGAN-GP appears to be fairly small, this could be further discouraged by incorporating the constraints of the human body into the loss function of the network. Therefore unusual motions would lead to an increased error, encouraging the network instead to produce more lifelike motion sequences.

In addition, as stated before, the frame sampling method we use means that the speed characteristics of a motion tend to be lost. Therefore, future work could involve learning a speed feature for each motion sequence. If this is learned, the feature could be used to rectify the motion speed during the rendering process.

Future work would involve a more thorough exploration of the motion space by the network, discovering how changes in the input noise correspond with changes in the characteristics of the outputted sample. Further work with regard to the conditioned model could also explore how positional information, such as start and end position of the bowl, can be used as a parameter to condition on. This increase in control over the output of the model may make the WGAN-GP model more suitable to applications in the animation industry. 

With regard to the motion classification task, in the same vein as Han et al.\cite{han}, future work could explore how the given object transport motions can be used to identify certain characteristics of the individual, such as age, or gender. 

In addition, besides comparing the results of the proposed convolutional model with a single cell LSTM, we should also compare our model's performance with that of more complex LSTM models on the dataset.

\appendix

\chapter{Additional WGAN-GP motion samples}

\begin{figure}[!htbp]
	\centering

		\makebox[\textwidth][c]{\includegraphics[width=1.35\textwidth]{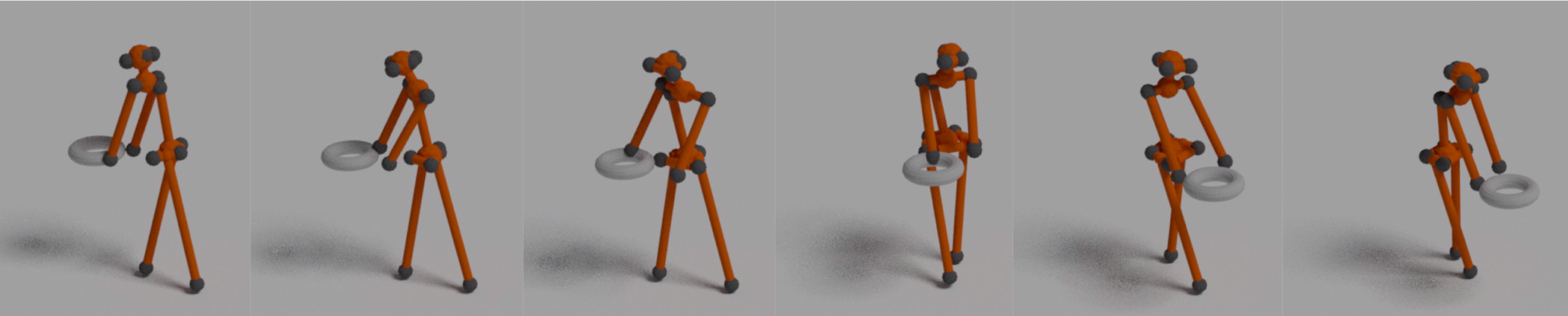}}

\end{figure}
\FloatBarrier
\begin{figure}[!htbp]
	\centering

		\makebox[\textwidth][c]{\includegraphics[width=1.35\textwidth]{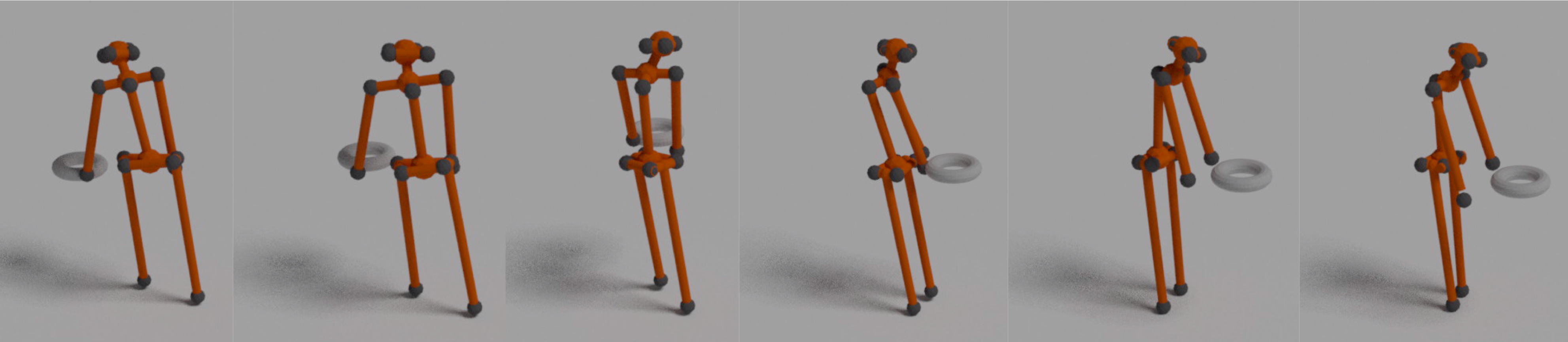}}

\end{figure}
\FloatBarrier
\begin{figure}[!htbp]
	\centering

		\makebox[\textwidth][c]{\includegraphics[width=1.35\textwidth]{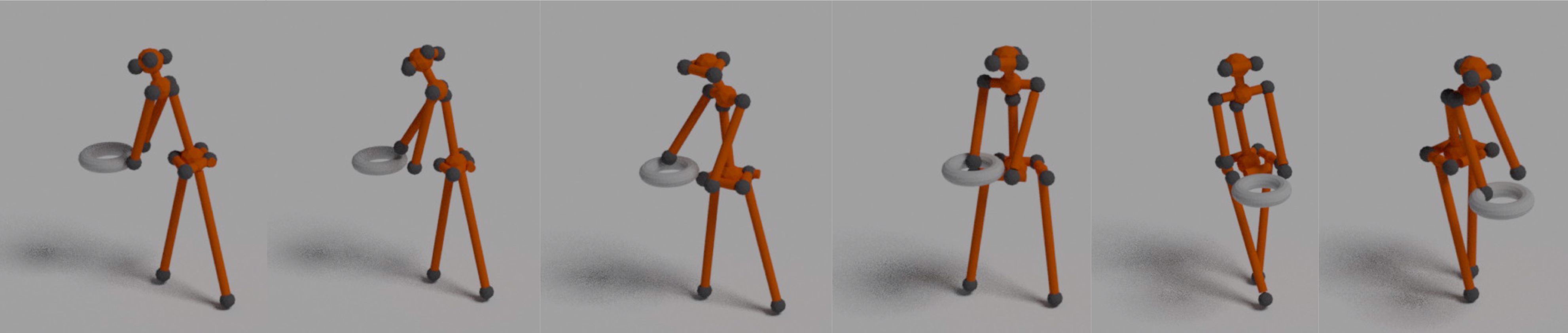}}

\end{figure}
\FloatBarrier
\begin{figure}[!htbp]
	\centering

		\makebox[\textwidth][c]{\includegraphics[width=1.35\textwidth]{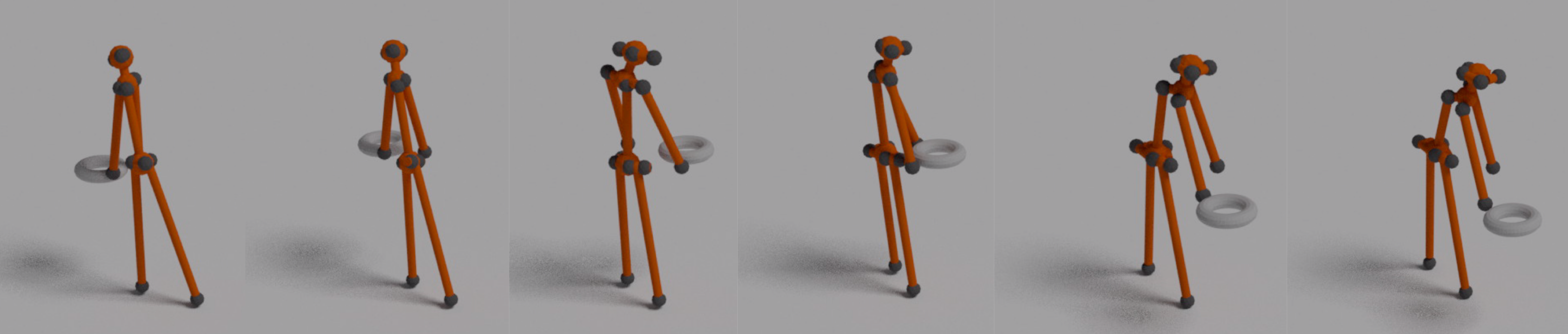}}

\end{figure}
\FloatBarrier
\begin{figure}[!htbp]
	\centering

		\makebox[\textwidth][c]{\includegraphics[width=1.35\textwidth]{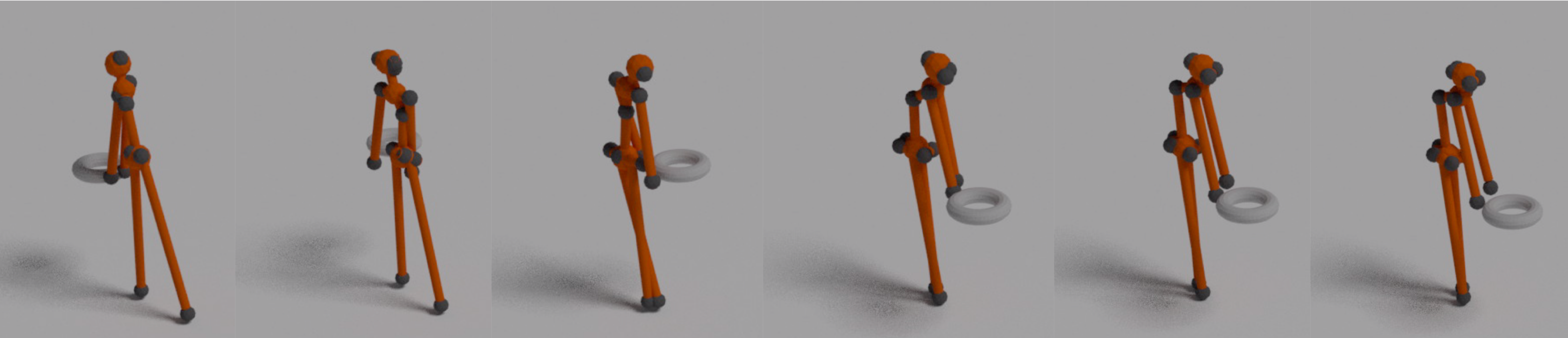}}

\end{figure}
\FloatBarrier

\chapter{Mid-project report}
\title{Generation of Object Transport Motion - Interim Report}

\author{Student: Connor Daly - Supervisors: Yuzuko Nakamura; Tobias Ritschel}

\date{January 2019}

\maketitle

\section{Generative Model - progress}
 DeepMind’s recent ‘Wavenet’ paper (2017) illustrates that Convolutional Neural Networks (CNNs), although typically used for image recognition, can be adapted for time sequential data and produce good results. Besides being easier to train than Long Short-Term Memory units (LSTMs), which are typically used for time sequential data, CNNs are also at the forefront of generative models. Deep Convolutional Generative Adversarial Networks, or DCGANs, are a type of generative adversarial network in which multilayer perceptrons are replaced with a convolutional network and a deconvolutional network. DCGANs were shown to outperform the original GAN network structure. Therefore, we decided to use DCGANs to generate novel examples of bowl transport motion from a dataset of motions. In order to create a DCGAN, we first needed to create a good CNN model suited to the given dataset. \newline

To help design this network, we set 3 dummy classification tasks of varying difficulty (based on how easy it was for a human observer to distinguish classes). The two easy tasks were: Determining if the object carried needed to be balanced (2 classes), and the strategy the participant was using to carry the object, such as purely left/right hand, both hands, hand-off etc. (9 classes). The more difficult task was determining the weight of the object being carried. In order to address an imbalance in proportion of weight classes in the data-set, this task was restricted to classifying between weights of either 640g or 1640g (2 classes) with bowls of the same size. \newline

Before passing the data to the networks, we shortened the length of the trial data by removing frames before and after the bowl was moved. We sampled every 12th frame of data (approximately 10 fps) until we had 32 frames (centred around the middle of the shortened data) giving an overall view of around 3.2 seconds of the trial when the bowl is in motion. The input data was then normalized. \newline

The data-set consisted of 874 trials. Given the comparatively small number of samples available to train and test with, it was difficult for the convolutional network to learn prior to over-fitting. Thus, we augmented the data. Augmentations were performed by shifting the coordinates along the X and Y axis, by scaling the participants body from a central point between the hip and shoulder markers and by rotating the motion paths from the bowl starting point. From these augmentations, we increased the size of the training data by a factor of 9. \newline

For the training process we used a 9:1 test train split.  For the first and second task (balanced-unbalanced, and strategy used) the model was able to classify well  (around 88\% after 400 epochs for task 1, and around 74\% after 400 epochs for task 2). For task 3, the smaller data-set available for training and testing (around 435 total samples compared to the original 874) meant that training was less stable. However with high dropout to assist regularization, we were able to achieve 70\% accuracy with the CNN. This was compared to a baseline of 65\% given by a single layer LSTM model. \newline

The final CNN model we used made use of 1-Dimensional dilated convolutions, as it allowed the model to learn across a greater number motion capture frames without increasing the parameters of the model. Markers were set into three clusters: head and shoulders, upper body and arms, lower body and legs. These clusters were then fed through parallel convolutions (3 layers), which were then concatenated and run through 2 final dense layers. 

\subsection{Rendering Generative Model Results - Progress}
We created a script within the 3D modelling package Blender, that reads in a CSV file of motion capture data, and generates a skeleton from the 15 body markers. We then generate cylindrical 'bones' to connect these points. The script then runs the animation of the participant moving the bowl object. This script will be re-purposed to render the generated motion results later in the project.

\section{Work to Complete}
Further work still needs to be done on fully understanding the nuances of designing and training DCGAN models. Using our results and model from the analyses tasks will hopefully expedite this process. We aim to incorporate elements from our current CNN design into the generator and discriminator network of the generative model. Generative models are notoriously difficult to train, consequently majority of the time between now and the deadline will be spent fine-tuning the model. Once the model is capable of generating motion capture data, we will be able to export it as a CSV file and render it accordingly in Blender.

\chapter{Project plan}

Project Supervisor: Yuzuko Nakamura
Project Author: Connor Daly

\title{Using deep learning to generate natural movement for a virtual actor}

\section{project outline}

This project is concerned with taking pre existing motion capture data taken by Yuzuko, and using it to generate motion plans for a virtual character.

The data, in its current form, examines the movement and rotation of the body when moving one object from one spot to another. We wish to use this data to generate a model detailing a realistic motion path for how a virtual actor would move an object across a room, given some high level input. The model would primarily be concerned with upper body movement and rotation, as research in this area is sparse.

Having examined several previous papers that examine motion generation with respect to gait, there are two main approaches in the field of motion generation: the first being physical modelling, which requires no input data, and relies entirely on analysis of the efficiency of the motion path with respect to the biomechanics of the virtual actor, whilst the second approach is data driven.

We are examining the second approach. Previous research from this side has made frequent use of advances in the field of deep learning. To this end, we will trial different neural networks and see how effectively they model the task at hand.

\section{timeline}

Nov 5-12
Get familiar with tensorflow and tflearn deep learning frameworks.
Explore different network architectures (eg. RNN, CNN etc.)

Nov 16-Dec 20
Begin design of network alongside Tobias Ritschel

Nov 5-12
Examine how to port motion capture into Blender environment

Dec 20 -Jan 15
Begin testing how network performs on current dataset. Current concern: lack of freedom with regard to placing of object in initial dataset may prove to be an obstacle. Decide whether further data needs to be collected.

Jan 10 - Jan 20
Once an accurate model has been generated, port data to blender and create skeleton for actor to be driven by generated motion paths. Look at using inverse kinematics to generate position of elbow.

Jan 20-Jan 30
Once model and actor are linked, begin qualitative feedback gathering.

Feb 1 onwards
Write up results in final paper

\end{document}